\documentclass[letterpaper]{article} % DO NOT CHANGE THIS
\usepackage{aaai25}  % !!!PUBLISH-CHECK!!! CAMERA READY: DO NOT CHANGE THIS
\nocopyright % !!!PUBLISH-CHECK!!! Arxiv version
\usepackage{times}  % DO NOT CHANGE THIS
\usepackage{helvet}  % DO NOT CHANGE THIS
\usepackage{courier}  % DO NOT CHANGE THIS
\usepackage[hyphens]{url}  % DO NOT CHANGE THIS
\usepackage{graphicx} % DO NOT CHANGE THIS
\usepackage{natbib}  % DO NOT CHANGE THIS AND DO NOT ADD ANY OPTIONS TO IT
\usepackage{caption} % DO NOT CHANGE THIS AND DO NOT ADD ANY OPTIONS TO IT
\usepackage{algorithm}
\usepackage{algorithmic}

\usepackage{mathtools}      % amsmath with fixes and additions
\usepackage{amssymb}        % Automatically loads amsfonts.
\usepackage{amsopn}
\usepackage{xparse}
\usepackage{cleveref}
    
\usepackage{newfloat}
\usepackage{listings}
\DeclareCaptionStyle{ruled}{labelfont=normalfont,labelsep=colon,strut=off} % DO NOT CHANGE THIS
\floatstyle{ruled}
\newfloat{listing}{tb}{lst}{}
\floatname{listing}{Listing}
\newtheorem{theorem}{Theorem}

\NewDocumentCommand \prArg{mm}
{(
\IfNoValueTF{#2}{#1}{#1 \mid #2}
)}

\NewDocumentCommand \newProbabilityFormat{r<>m}
{
	\NewDocumentCommand #1 {e{_}e{^}>{\SplitArgument{1}{|}}d()}
	{
		\IfNoValueTF{##1}
		{
			\IfNoValueTF{##2}
			{\IfNoValueTF{##3}{#2}{#2\prArg##3}}
			{\IfNoValueTF{##3}{#2^{##2}}{#2^{##2}\prArg##3}}
		}
		{
			\IfNoValueTF{##2}
			{\IfNoValueTF{##3}{#2_{##1}}{#2_{##1}\prArg##3}}
			{\IfNoValueTF{##3}{#2_{##1}^{##2}}{#2_{##1}^{##2}\prArg##3}}
		}
	}
}

\NewDocumentCommand \fVector {m} {\boldsymbol{#1}}
\NewDocumentCommand \fMatrix {m} {\boldsymbol{#1}}

\NewDocumentCommand \fFunction {m} {{#1}}
\NewDocumentCommand \fSet {m} {\mathcal{#1}}

\NewDocumentCommand \fOperator {m} {\mathop{\vphantom{\sum}\mathchoice {\vcenter{\hbox{\Large $#1$}}} {\vcenter{\hbox{\Large $#1$}}}{#1}{#1}}\displaylimits}

\NewDocumentCommand \newScalar{r<>m}
{
	\NewDocumentCommand #1 {} {{#2}}
}

\NewDocumentCommand \newVector{r<>m}
{
	\NewDocumentCommand #1 {} {\fVector{#2}}
}

\NewDocumentCommand \newMatrix{r<>m}
{
	\NewDocumentCommand #1 {} {\fMatrix{#2}}
}

\NewDocumentCommand \newProbability{r<>m}
{
	\newProbabilityFormat<#1>{#2}
}

\NewDocumentCommand \newFunction{r<>m}
{
	\NewDocumentCommand #1 {} {\fFunction{#2}}
}

\NewDocumentCommand \newSet{r<>m}
{
	\NewDocumentCommand #1 {} {\fSet{#2}}
}

\NewDocumentCommand \oH {} {\fOperator{\ast}} % Hadamard, elementwise
\DeclareMathOperator{\expect}{\mathbb{E}}

\DeclareMathOperator{\diag}{diag}

\DeclareMathOperator{\sign}{sign}

\NewDocumentCommand \reals {} {\mathbb{R}}

\NewDocumentCommand \qed {} {{\:}_\blacksquare}

\newScalar<\eps>{\varepsilon}

\newVector<\vA>{a}
\newVector<\vB>{b}
\newVector<\vC>{c}
\newVector<\vD>{d}
\newVector<\vE>{e}
\newVector<\vF>{f}
\newVector<\vG>{g}
\newVector<\vH>{h}
\newVector<\vI>{i}
\newVector<\vJ>{j}
\newVector<\vK>{k}
\newVector<\vL>{\ell}
\newVector<\vM>{m}
\newVector<\vN>{n}
\newVector<\vP>{p}
\newVector<\vQ>{q}
\newVector<\vR>{r}
\newVector<\vS>{s}
\newVector<\vT>{t}
\newVector<\vU>{u}
\newVector<\vV>{v}
\newVector<\vW>{w}
\newVector<\vX>{x}
\newVector<\vY>{y}
\newVector<\vZ>{z}
\newVector<\vAlpha>{\alpha}
\newVector<\vBeta>{\beta}
\newVector<\vGamma>{\gamma}
\newVector<\vDelta>{\delta}
\newVector<\vEpsilon>{\varepsilon}
\newVector<\vEps>{\varepsilon}
\newVector<\vZeta>{\zeta}
\newVector<\vEta>{\eta}
\newVector<\vTheta>{\theta}
\newVector<\vTh>{\theta}
\newVector<\vKappa>{\kappa}
\newVector<\vLambda>{\lambda}
\newVector<\vMu>{\mu}
\newVector<\vNu>{\nu}
\newVector<\vXi>{\xi}
\newVector<\vPi>{\pi}
\newVector<\vRho>{\rho}
\newVector<\vSigma>{\sigma}
\newVector<\vTau>{\tau}
\newVector<\vUpsilon>{\upsilon}
\newVector<\vPhi>{\varphi}
\newVector<\vChi>{\chi}
\newVector<\vPsi>{\psi}
\newVector<\vOmega>{\omega}

\newMatrix<\mA>{A}
\newMatrix<\mB>{B}
\newMatrix<\mC>{C}
\newMatrix<\mD>{D}
\newMatrix<\mE>{E}
\newMatrix<\mF>{F}
\newMatrix<\mG>{G}
\newMatrix<\mH>{H}
\newMatrix<\mI>{I}
\newMatrix<\mJ>{J}
\newMatrix<\mK>{K}
\newMatrix<\mL>{L}
\newMatrix<\mM>{M}
\newMatrix<\mN>{N}
\newMatrix<\mP>{P}
\newMatrix<\mQ>{Q}
\newMatrix<\mR>{R}
\newMatrix<\mS>{S}
\newMatrix<\mT>{T}
\newMatrix<\mU>{U}
\newMatrix<\mV>{V}
\newMatrix<\mW>{W}
\newMatrix<\mX>{X}
\newMatrix<\mY>{Y}
\newMatrix<\mZ>{Z}
\newMatrix<\mGamma>{\Gamma}
\newMatrix<\mDelta>{\Delta}
\newMatrix<\mTheta>{\Theta}
\newMatrix<\mLambda>{\Lambda}
\newMatrix<\mXi>{\Xi}
\newMatrix<\mPi>{\Pi}
\newMatrix<\mSigma>{\Sigma}
\newMatrix<\mUpsilon>{\Upsilon}
\newMatrix<\mPhi>{\Phi}
\newMatrix<\mPsi>{\Psi}
\newMatrix<\mOmega>{\Omega}

\newSet<\sA>{A}
\newSet<\sB>{B}
\newSet<\sC>{C}
\newSet<\sD>{D}
\newSet<\sE>{E}
\newSet<\sF>{F}
\newSet<\sG>{G}
\newSet<\sH>{H}
\newSet<\sI>{I}
\newSet<\sJ>{J}
\newSet<\sK>{K}
\newSet<\sL>{L}
\newSet<\sM>{M}
\newSet<\sN>{N}
\newSet<\sP>{P}
\newSet<\sQ>{Q}
\newSet<\sR>{R}
\newSet<\sS>{S}
\newSet<\sT>{T}
\newSet<\sU>{U}
\newSet<\sV>{V}
\newSet<\sW>{W}
\newSet<\sX>{X}
\newSet<\sY>{Y}
\newSet<\sZ>{Z}

\title{Curvature in the Looking-Glass: Optimal Methods to\\Exploit Curvature of Expectation in the Loss Landscape}
\author{
    Jed A.~Duersch\textsuperscript{\rm 1},
		Tommie A.~Catanach\textsuperscript{\rm 1},
		Alexander Safonov\textsuperscript{\rm 1},
		Jeremy Wendt\textsuperscript{\rm 2}
}
\affiliations{
    \textsuperscript{\rm 1}Sandia National Laboratories, Livermore, CA, USA\\
		\textsuperscript{\rm 2}Sandia National Laboratories, Albuquerque, NM, USA\\
    \{jaduers, tacatan, amsafon, jdwendt\}@sandia.gov
}

\begin{document}

\maketitle

% !!! REMOVE xcolor BEFORE SUBMISSION !!!
%\NewDocumentCommand \anonymize{O{}m} {#1} % !!!PUBLISH-CHECK!!! Anonymized
\NewDocumentCommand \anonymize{O{}m} {#2} % !!!PUBLISH-CHECK!!! Camera-ready
%\NewDocumentCommand \jNote{m} {{\color{red}#1}}

\NewDocumentCommand \appendixB{} {Appendix B}
\NewDocumentCommand \appendixC{} {Appendix C}

\begin{abstract} %237
Harnessing the local topography of the loss landscape is a central challenge in advanced optimization tasks.
By accounting for the effect of potential parameter changes, we can alter the model more efficiently.
Contrary to standard assumptions, we find that the Hessian does not always approximate loss curvature well,
particularly near gradient discontinuities, which commonly arise in deep learning architectures.

We present a new conceptual framework to understand how curvature of expected changes in loss
emerges in architectures with many rectified linear units.
Each ReLU creates a parameter boundary that, when crossed, induces a pseudorandom gradient perturbation.
Our derivations show how these discontinuities combine to form a glass-like structure,
similar to amorphous solids that contain microscopic domains of strong, but random, atomic alignment.
By estimating the density of the resulting gradient variations, we can bound how the loss may change with parameter movement.
Our analysis includes the optimal kernel and sample distribution for approximating glass density from ordinary gradient evaluations.
We also derive the optimal modification to quasi-Newton steps that incorporate both glass and Hessian terms,
as well as certain exactness properties that are possible with Nesterov-accelerated gradient updates.

Our algorithm,
Alice%
%\footnote{\url{https://github.com/sandialabs/alice}}% % !!!PUBLISH-CHECK!!! Potential alternative to AAAI copyright.
, tests these techniques to determine which curvature terms are most impactful for training a given architecture and dataset.
Additional safeguards enforce stable exploitation through step bounds that expand on the functionality of Adam.
These theoretical and experimental tools lay groundwork to improve future efforts (e.g., pruning and quantization)
by providing new insight into the loss landscape.
\end{abstract}

% \anonymize%
% [\link{Code}{Anonymized.}]%
% {\link{Code}{https://github.com/sandialabs/alice}}
% % \link{Extended version}{https://aaai.org/example/extended-version}

% =========================================================================================================================
% =========================================================================================================================
% =========================================================================================================================
% ===================================================== DEFINITIONS =======================================================
% =========================================================================================================================
% =========================================================================================================================
% =========================================================================================================================

% SCALARS
\newScalar<\dist>{\lambda}
\newScalar<\reluThres>{\psi}
\newScalar<\dropThres>{\lambda_{\rm min}}
\newScalar<\nPa>{d}
\newScalar<\nSa>{s}
\newScalar<\krnCnst>{c}
\newScalar<\lrMin>{\lambda_{\rm min}}
\newScalar<\lrMax>{\lambda_{\rm max}}
\newScalar<\variance>{\rm V}

% FUNCTIONS
\newFunction<\fLss>{\mathcal{L}}
\newFunction<\fGrd>{g}
\newFunction<\fHss>{H}
\newFunction<\fRelu>{{\rm ReLU}}
\newFunction<\fKrn>{\kappa}
\newFunction<\fKrnPrt>{\eta}

% VECTORS
\newVector<\vPa>{\theta}
\newVector<\vPaCnt>{\mu}
\newVector<\vPaEvl>{\nu}
\newVector<\vPaPrt>{\delta}

\newVector<\vGrd>{g}
\newVector<\vGrdPrt>{\gamma}
\newVector<\vGrdY>{\hat{\gamma}}
\newVector<\vGrdVar>{v}
\newVector<\vGrdVarDns>{\rho}

\newVector<\vHss>{h}
\newVector<\vHssGls>{\hat{h}}
\newVector<\vHssMod>{\bar{h}}
\newVector<\vMatDiag>{m}
\newVector<\vSec>{s}
\newVector<\vHssRms>{h_{\rm rms}}

% MATRICES
\newMatrix<\mHss>{H}
\newMatrix<\mGrdVarDns>{R}
\newMatrix<\mMat>{M}

% SETS
\newSet<\sPaPart>{P}
\newSet<\sRelu>{S}

% DENSITIES
\newProbability<\pN>{\mathcal{N}}
\newProbability<\pP>{p}
\newProbability<\pR>{r}
\newProbability<\pDelta>{\delta_{\rm D}}
\newProbability<\pPrt>{\eta}

\newFunction<\dL>{\Delta}
\newFunction<\dLoss>{\Delta\mathcal{L}}

% =================================================================
% ========================= Introduction ==========================
% =================================================================
\section{Introduction}

We present a mathematical framework to understand how distributed gradient discontinuities arise in neural networks
and the effect this phenomenon has on the topography of the loss landscape.
This framework opens new optimization methods to estimate and exploit these effects.

This work began with an effort to derive the optimal kernel for estimating locally-averaged linear dependence of the gradient on parameters.
During testing, however, the following experiment produced unexpected results.
As a consequence, we find that the Hessian is not always % the dominant source of curvature of expectation in the loss landscape.
the next term, after the gradient, needed to approximate loss topography.

Let $\fLss(\vPa)$ represent average training loss, with gradient $\fGrd(\vPa)$, for a network with parameters $\vPa \in \reals^\nPa$.
We seek to understand how gradient changes depend on the scale of model perturbations
to confirm whether a Hessian $\mHss$ dominates.
Centering parameters at $\vPa = \vPaCnt$, we draw a vector of i.i.d.~Rademacher samples $\vPaPrt$ (i.e.~each element is equally likely to be $\pm1$)
and then measure average gradient variations $\vGrdVar(\dist)$ corresponding to a scalar distance $\dist$ as
\begin{align}
\label{eq:grd_var}
\vGrdVar(\dist) = \expect\left[ \vGrdPrt^2 \right]
\quad\text{where}\quad
\vGrdPrt = \fGrd(\vPaCnt + \dist \vPaPrt) - \fGrd(\vPaCnt).
\end{align}
Here, and throughout this paper, powers of vectors are applied elementwise.
We use the term \textit{gradient variations} to distinguish from variance (centered at the mean).
This yields a vector of second moments centered at $\fGrd(\vPaCnt)$.
By measuring gradient variations at both $\dist$ and $2\dist$, we can approximate a power law, $\vGrdVar_i(\dist) \approx \vK_i \dist^p$, over any parameter subset $\sPaPart$ as
\begin{align}
\label{eq:power_law}
p(\sPaPart) = \log_2 \sum_{i \in \sPaPart} \vGrdVar_i(2 \dist) - \log_2 \sum_{i \in \sPaPart} \vGrdVar_i(\dist).
\end{align}

\begin{figure}[t]
\centering
\includegraphics[width=0.98\columnwidth]{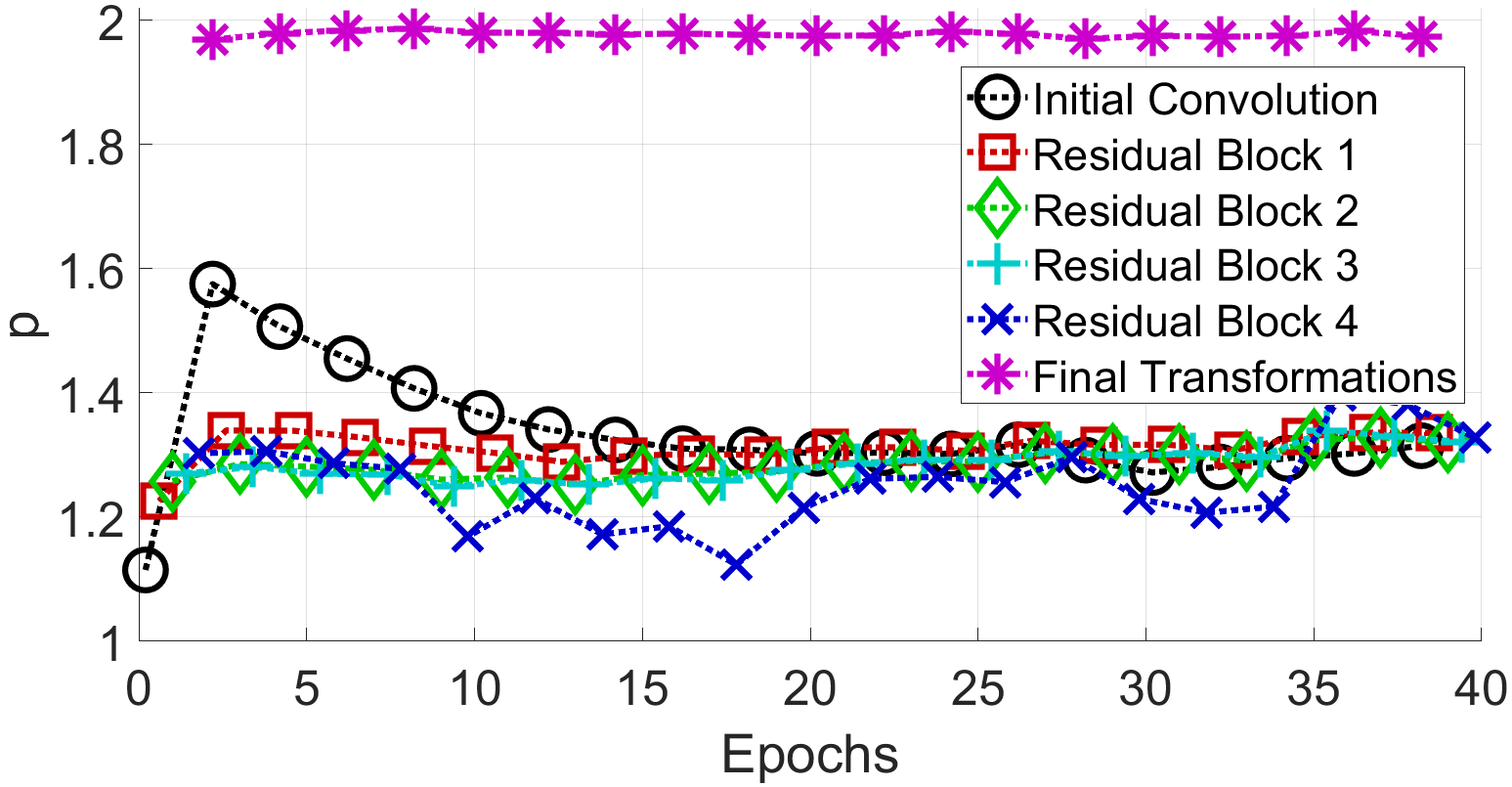} 
\caption{Exponential dependence of gradient variations on perturbations from \Cref{eq:power_law}.
A Hessian term yields $p = 2$.
No low-order Taylor series can give $p < 2$.}
%Power laws, $\vGrdVar_i(\dist) \approx \vK_i \dist^p$, approximating how gradient variations depend on distance in ResNet18.}
\label{fig:grad_var_exp}
\end{figure}

If the gradient depended linearly on perturbations, i.e.~$\vGrdPrt = \dist \mHss \vPaPrt$, then \Cref{eq:power_law} would give $p=2$.
%A large second-order term could push $p$ up to $4$.
\Cref{fig:grad_var_exp} shows $p$ for ResNet18 \citep{he2016deep} with CIFAR-10 \citep{Krizhevsky2009} over 40 epochs with $\dist = 0.002$.
Parameters are partitioned by: 1.~initial convolutions, 2-5.~each residual block, and 6.~final transformations.

These results cannot be explained by a low-order truncated Taylor series, for which $p \geq 2$.
%The particular values of $p$ are not as important as the evidence of a balance between linear and quadratic terms (i.e. sublinear and linear, respectively, in the gradient).
Rather, these values of $p$ suggest a balance between linear and quadratic terms in gradient variations (i.e. sublinear and linear, respectively, in the gradient).
\textit{Thus, a coherent mathematical framework to explain this phenomenon is needed.}

\begin{figure}[t]
\centering
\includegraphics[width=0.9\columnwidth]{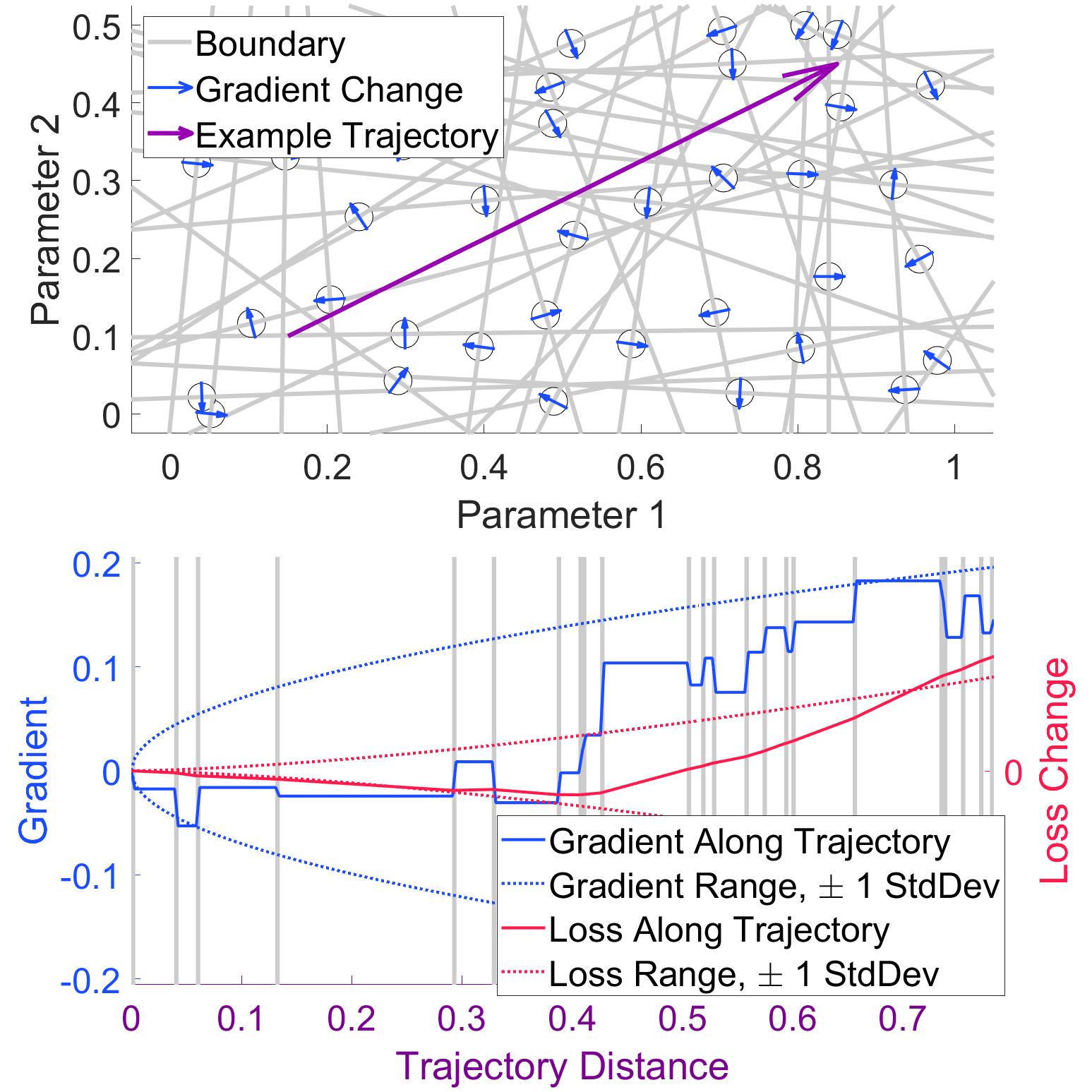} % Slightly narrower than the column.
% Captions must appear under the illustration. Text inside actual image must be at least nine point type!
% Do not make font size in captions smaller, bold, or italic (other than individual words)!
% Use of color is restricted to figures. Be mindful of readers have trouble distinguishing colors.
\caption{Illustration of gradient glass in 2D (top).
Gray lines are domain boundaries due to changing a ReLU state.
Blue arrows show the resulting gradient perturbations.
We can compute the curvature of expected changes (bottom) to the gradient and loss from the density of these variations.}
\label{fig:glass_alt}
\end{figure}

% ===============================
% ======== Contributions ========
% ===============================
\paragraph{Contributions}
We present an analytic framework for loss topography based upon a gradient glass structure
that recovers linearity in gradient variations.
The term \textit{glass} alludes to the structure of an amorphous solid,
comprising small domains of strong molecular alignment that randomly shifts at domain boundaries.
\Cref{fig:glass_alt} illustrates a gradient glass in 2D (top)
and the resulting loss effects, including expectation bounds (bottom).
Crossing a domain boundary (gray line) results in a pseudorandom addition to the gradient (blue arrows).
We can derive this structure from ReLUs in the network \citep{nair2010rectified}.
The results in \Cref{fig:grad_var_exp} are also consistent with this explanation.
Since the final parameters are not followed by ReLUs, their gradients do not encounter glass discontinuities, leaving $p=2$ from the Hessian.

Despite the jagged effects of a gradient glass on the loss topography,
we can still analyze and compute the \textit{curvature of expectation}, including the contribution from a locally averaged Hessian, to improve loss extrapolation for various optimization tasks.
Our derivations include:
\begin{enumerate}
\item the density of gradient variations from ReLUs,
\item the optimal kernel for estimating diagonal dependence of both the locally-averaged Hessian and glass density,
\item the optimal corresponding sample distribution,
\item upper bounds on expected changes in loss,
\item the optimal modification to quasi-Newton steps, and
\item exactness properties from Nesterov acceleration.
\end{enumerate}

Although our primary contributions are theoretical, basic loss reduction is a necessary stage of development
before expanding to other optimization tasks.
Thus, we present a simple training algorithm, Alice%
\footnote{\anonymize[[URL of source code] (Anonymized for peer-review.)]{\url{https://github.com/sandialabs/alice}}}% !!!PUBLISH-CHECK!!!
% Potentially remove for arxiv, anonymous for peer-review, url for camera ready.
, to examine these techniques in practice and provide a baseline for further advances that leverage glass-based curvature.
While well-tuned learning schedules can eventually converge to models with strong performance,
rapid loss reduction can only be achieved with high-quality topographical information.
Thus, Alice can reveal which curvature terms and length scales are useful for understanding a curious landscape.

\Cref{sec:related} discusses related work
followed by our analytic results in \Cref{sec:derivations}.
\Cref{sec:methods} provides the implementation details used in Alice
along with numerical experiments.
\Cref{sec:discussion} provides %a brief discussion of key points of interest and concluding remarks.
concluding remarks.
Proofs are provided in \Cref{sec:proofs}.
The extended version of this paper contains additional methodological details (\appendixB) and experimental setups (\appendixC).

% ===================================================================
% =========================== Related Work ==========================
% ===================================================================
\section{Related Work}
\label{sec:related}

Quasi-Newton (QN) methods use an invertible Hessian approximation to capture local curvature
and attempt more efficient parameter updates, % that resemble exact Newton steps,
if the loss is smooth enough for the Hessian to be informative.
Work incorporating curvature into ML optimization, via second-order terms, began with \citet{becker1988improving}.
In multilayer perceptrons, the Hessian diagonal $\vHss(\vPa)$ is directly computed by propagating second derivatives on each hidden unit.
This facilitates pruning \citep{lecun1989optimal}
or improves convergence with QN steps,
$\vPaCnt \gets \vPaCnt - \fGrd(\vPaCnt) \oH \left( | \vHss(\vPaCnt) | + \eps \right)^{-1}$.
We denote the Hadamard product with $\oH$.
%\citet{lecun1990second} also show how the spectrum of the full Hessian informs learnability.

Since then, there has been extensive work on second-order methods.
\Citet{bottou2018optimization} provide an overview of these developments.
%For example, \citet{agarwal2016second} approximate Newton steps by approximating the Hessian inverse as a product of identity plus low-rank matrices.
%For example, Hessian-free inexact Newton methods exploit a conjugate gradient formulation of the inverse that only requires Hessian matrix-vector products.
Notably, they explain how the Hessian need not be as exact as the gradient, allowing for sub-sampled estimators (updating Hessian information less frequently or with smaller batches) to speed up optimization.

Recently, \citet{yao2020pyhessian} present methods to probe the spectrum of the Hessian for large networks. 
They avoid forming an explicit Hessian by using a second stage of backpropagation to obtain Hessian-vector products.
If $\vX$ is a given vector in the product $\fHss(\vPa) \vX$, then one computes $\nabla_{\vPa} \fGrd(\vPa)^T \vX$ after the first stage of backpropagation. %, since
%\begin{displaymath}
%\fHss_{ij} = \frac{\partial^2 \fLss(\vPa)}{\partial \vPa_i \partial \vPa_j}
%\quad\text{so}\quad
%\frac{\partial \sum_j \fGrd_j(\vPa) x_j}{\partial \vPa_i} =  \sum_j \fHss_{ij} x_j.
%\end{displaymath}
Using power iteration and randomized methods, their tools provide insight into how architecture design affects optimization.

\paragraph{Active Developments}
The following summaries, by no means extensive, indicate the active interest in using curvature to advance optimization, due in part to the potential connection with generalizability.
\Citet{lyle2023understanding} examine how plasticity (a model's ability to absorb new data) is heavily influenced by changes in loss curvature,
and \Citet{kong2023overcoming} shows that catastrophic forgetting can be mitigated by controlling the maximum Hessian eigenvalue.
In transfer learning, \Citet{hemati2023understanding} forge connections between the Hessian, specifically at the classification head, and out-of-distribution transfer.
\Citet{li2023understanding} investigate whether curvature-based regularization can benefit adversarial training objectives.
To exploit curvature,
\Citet{sen2023federated} use diagonal Hessian approximations to improve convergence of federated, data-parallel, training.
\Citet{kaur2023maximum} analyze the relationship between curvature and generalization,
showing how that the largest Hessian eigenvalues can be controlled through batch size and learning rate adjustments in Sharpness-Aware Minimization (SAM).
\Citet{lee2023shot} also present an algorithm to suppress Hessian curvature along optimization pathways,
and \Citet{lee2023achieving} use learning rate adjustments to encourage convergence at flat minima.
\citet{duersch2024projective} provides a method to merge QN with variational inference,
showing how the Hessian diagonal can calibrate model uncertainty while improving parameter updates.

Our derivations also include a variation of Nesterov accelerated quasi-Newton (NAQ),
initially proposed by \Citet{ninomiya2017novel} and \citet{indrapriyadarsini2020stochastic}.
They combine QN methods with Nesterov acceleration, focusing on the symmetric low-rank Hessian updates of Broyden-Fletcher-Goldfarb-Shanno (BFGS) methods.
The cost of these methods scales linearly with approximation rank, which can be prohibitive for very large architectures.
See \cite{nocedal2006theory} for details regarding BFGS.

\paragraph{Hessian Shortcomings}
Despite the many successes of Hessian-based approaches,
the resulting interpolants are not always the best way to approximate local curvature,
particularly in the vicinity of gradient discontinuities.

\begin{figure}[t]
\centering
\includegraphics[width=0.9\columnwidth]{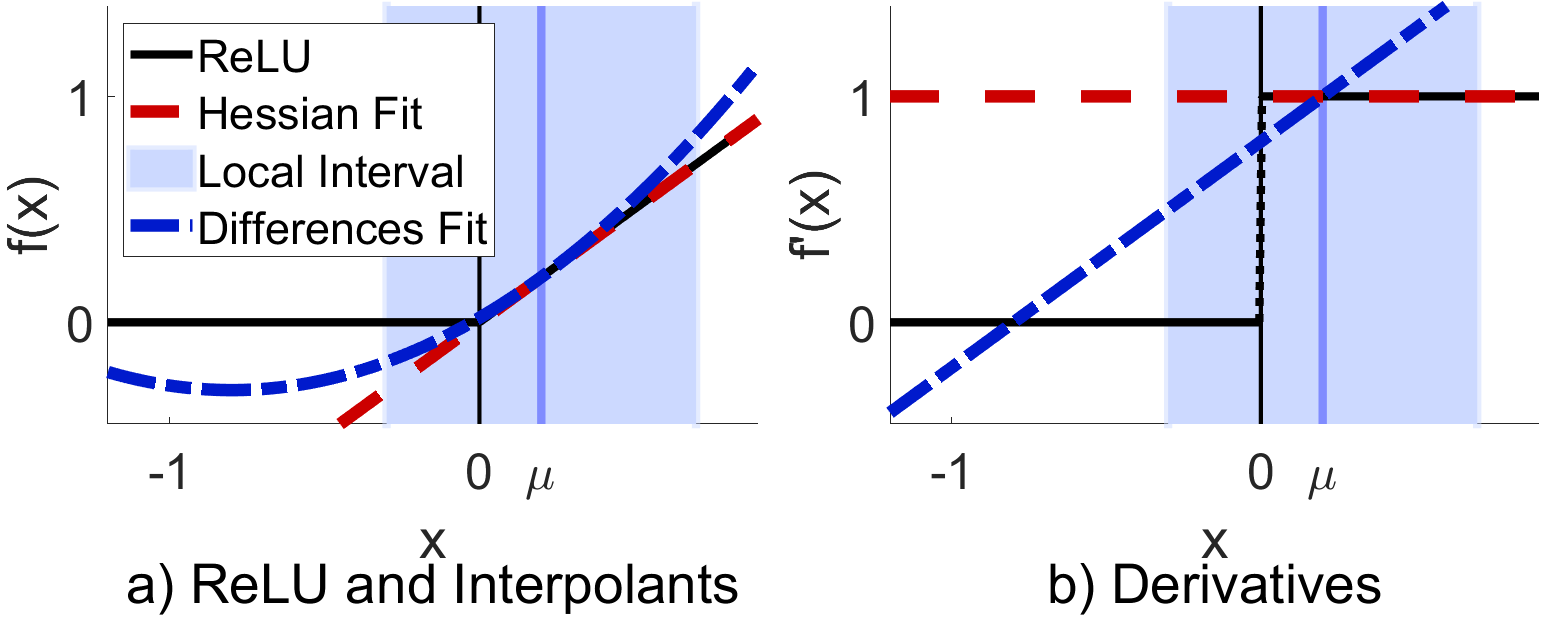} % Slightly narrower than the column.
\caption{ReLU extrapolation. Hessian fitting cannot account for gradient discontinuities, whereas curvature matching gradient changes provides a better fit on a local interval.}
\label{fig:relu_interp}
\end{figure}

\Cref{fig:relu_interp} illustrates this issue with a basic ReLU, $f(x) = \max(0, x)$.
Curvature matching the difference of gradients follows local changes and anticipates gradient zeros,
whereas the Hessian fails to obtain any curvature that could inform optimization.
Work by \Citet{frumkin2023jumping}, regarding how to account for the jagged loss topography they observe during quantization,
also supports this picture of loss topography.
Moreover, \Citet{li2024visualizing} find that loss typically has v-shaped structure along random trajectories, which bears a critical connection to a predicate in our \Cref{thm:glass_loss} (See \Cref{sec:derivations}), that random loss trajectories are reflected at a local floor.

\paragraph{Glass in the Literature}
Glasses are technically physical systems with non-convex energy functions of state.
There have been a few papers referencing a notion of a glass with respect to high-dimensional optimization problems.
\Citet{dauphin2014identifying} provide both analytic and experimental evidence for a proliferation of saddle points, rather than local minima, as model dimensionality increases.
This perspective draws insight from spin-glasses in statistical physics \citep{parisi2007course}.
Saddle points are critical points (gradient zeros) where $\mH$ contains both positive and negative eigenvalues.
In the vicinity of such points, low apparent curvature creates an illusion of convergence, despite the hidden potential to make progress.
To address this problem, they support saddle-free Newton, which repels saddle points by using $|\mH|$, the rectified matrix constructed from absolute values of Hessian eigenvalues \citep{nocedal2006theory}.

\Citet{choromanska2015loss} go on to show that, while many local minima may also exist, large networks provide high probability of convergence to solutions of similar performance,
with local minima of poor quality becoming increasingly rare.
\Citet{spigler2019jamming} also estabilish a connection between glasses and the phase transition in generalizability as deep networks change from under- to over-parameterization.
They show how the phenomenon of overfitting is most problematic in the vicinity of this transition.

The distribution of critical points in the loss topography directly arises from the gradient variations that we scrutinize.
Our objective is to lay groundwork for simple and computationally efficient methods to navigate the resulting topography to support novel optimization tasks.

% ==================================================================
% =========================== Derivations ==========================
% ==================================================================
\section{Derivations}
\label{sec:derivations}

%\Cref{thm:relu_pert} shows how the discontinuity in the derivative of a ReLU induces a gradient discontinuity in the loss.
\Cref{thm:relu_pert} relates discontinuities in ReLU derivatives to the loss' gradient discontinuities.
See \Cref{sec:proofs} for proofs.
%The boundary of the discontinuity is locally approximated by a hyperplane separating half-spaces. 
%The boundary, locally approximated by a hyperplane separating half-spaces, has a normal vector that is parallel to the resulting discontinuity.

\begin{theorem}[Glass from ReLUs]
\label{thm:relu_pert}
Consider a network with many ReLUs.
Let a single unit be $z = \max(y,0) \in \reals$.
To examine the effect of a small parameter perturbation, $\vPa = \vPaCnt + \vPaPrt$,
on the gradient variations defined by \Cref{eq:grd_var},
we hold network inputs fixed and assume each ReLU input is locally linearly dependent on the perturbation,
\begin{align}
\label{eq:y_expand}
y(\vPaCnt + \vPaPrt) = y(\vPaCnt) + \vPaPrt^T \vGrdY
\quad\text{where}\quad
\vGrdY = \nabla_{\vPaCnt} y(\vPaCnt).
\end{align}
Suppose a subset of ReLUs have pre-activation values that are uniformly distributed at random in an interval $[-\reluThres, \reluThres]$,
indexed by $\sRelu_{\reluThres} = \left\{ k \mid z^{(k)} = \max(y^{(k)},0),\: |y^{(k)}| < \reluThres \right\}$,
and perturbations satisfy $|\vPaPrt^T \vGrdY^{(k)}| < \reluThres$.
Then, treating the pre-activation gradients $\vGrdY^{(k)}$ as independent zero-mean pseudorandom variables,
the gradient variations are bound by $\vGrdVar(\vPaPrt) \leq \mGrdVarDns |\vPaPrt|$
where $\mGrdVarDns$ is a matrix % capturing the density of gradient variations with elements
with elements
\begin{align}
\mGrdVarDns_{ij} = \frac{1}{2 \reluThres} \sum_{k \in \sRelu_{\reluThres}} \vGrdY_i^{(k)2} \left(\frac{d\fLss(\vPaCnt)}{dz^{(k)}}\right)^2 |\vGrdY_j^{(k)}|.
\end{align}
\end{theorem}

\newcommand{\proofone}{
\paragraph{Proof of \Cref{thm:relu_pert}}
W.l.o.g., consider an inactive unit at $\vPa = \vPaCnt$, i.e.~$y(\vPaCnt) < 0$ so that $z(\vPaCnt) = 0$.
Using \Cref{eq:y_expand} to compute the perturbation needed to cross the activation threshold
gives $\vPaPrt^T \vGrdY \geq -y$.
This changes $\frac{\partial z}{\partial y}$ from $0$ to $1$,
so the chain rule gives a gradient perturbation $\vGrdPrt \approx \frac{d\fLss(\vPaCnt)}{dz} \vGrdY$ in \Cref{eq:grd_var}.
Crossing in reverse yields a minus sign.
Thus, each ReLU generates a half-space containing an additive gradient perturbation.
When $\vPaPrt$ crosses the threshold (in either direction), $\vGrdPrt^2$ is added to the gradient variations.

We only need to consider these effects for ReLUs in $\sRelu_{\reluThres}$, which have small enough inputs to potentially flip.
Since $y^{(k)}$ are distributed uniformly at random over $[-\reluThres, \reluThres]$ for $k \in \sRelu_{\reluThres}$,
the probability that a specific $\vPaPrt$ is in the correct direction and large enough to cross the threshold is
$\pP\!\left( -\sign(y^{(k)}) \vPaPrt^T \vGrdY^{(k)} > |y^{(k)}| \right) = \frac{|\vPaPrt^T \vGrdY^{(k)}|}{2 \reluThres}$.

Treating these $\vGrdY$ as independent mean zero pseudorandom variables,
$\vGrdVar$ is the diagonal of the variance of the sum over all resulting $\vGrdPrt$ perturbations,
which is just the sum of individual variances.
Using the triangle inequality, $|\vPaPrt^T \vGrdY^{(k)}| \leq \sum_{j=1}^{\nPa} |\vGrdY_j^{(k)} \vPaPrt_j|$,
the total variance is
\begin{displaymath}
\vGrdVar(\vPaPrt) \leq \frac{1}{2 \reluThres}\!\sum_{k \in \sRelu_{\reluThres}}\! \vGrdY^{(k)2} \left( \frac{d\fLss(\vPaCnt)}{dz^{(k)}} \right)^{2} \sum_{j=1}^{\nPa} \left| \vGrdY_j^{(k)} \right| \left| \vPaPrt_j \right|\qed
\end{displaymath}
}
%\proofone

We interpret $\mGrdVarDns$ as a density matrix that captures additional gradient variations per unit length.
Any activation function with ReLU-like discontinuities in the derivative will induce discontinuities in the loss gradient.
The same reasoning holds for activation functions with rapidly changing derivatives that can be approximated by a step function.
Although \Cref{thm:relu_pert} is restricted to a fixed input,
if we consider a full dataset,
averaging diminishes the size of individual discontinuities, but they still remain.
Critically, point-wise Hessian methods cannot detect their presence or effects.% since they only appear from specific parameter changes.

Note the dependence of $\mGrdVarDns_{ii}$ on $|\vGrdY_i^{(k)}|^3$.
The largest magnitude element, $|\vGrdPrt_{i_{\text{max}}}^{(k)}| > |\vGrdPrt_j^{(k)}|$ for all $j \neq i_{\text{max}}$,
has a much stronger diagonal effect,
$| \vGrdPrt_{i_{\text{max}}}^{(k)} |^3 > | \vGrdPrt_{i_{\text{max}}}^{(k)} |^2 | \vGrdPrt_j^{(k)} |
> | \vGrdPrt_j^{(k)} |^2 |\vGrdPrt_{i_{\text{max}}}^{(k)}|$.
Since the largest element differs over the various $k \in \sRelu_{\reluThres}$,
these dominant terms are distributed over many diagonal elements.
Thus, taking $\vGrdVar \approx \vGrdVarDns \oH |\vPaPrt|$, where $\vGrdVarDns = \diag(\mGrdVarDns)$,
is a simple and reasonable approximation.

\Cref{thm:optimal_kernel} estimates the diagonal of linear operators by sampling matrix-vector multiplies,
allowing both $\vGrdVarDns$ and $\vHss$, the diagonal of the locally averaged Hessian,
to be computed from ordinary gradients evaluated at sample locations.

\begin{theorem}[Optimal Kernel for Estimating Diagonal]
\label{thm:optimal_kernel}
We are given a probability density over perturbations $\pP(\vPaPrt)$ as a product of i.i.d.~factors,
$\pP(\vPaPrt) = \prod_{i=1}^\nPa \pP(\vPaPrt_i)$,
each having zero mean and unit variance.
If we have a mechanism to evaluate matrix-vector products, $\vY = \mMat \vPaPrt$,
then we can estimate each element $\vMatDiag_i$ of the diagonal $\vMatDiag = \diag(\mMat)$
using samples weighted by a kernel function of the corresponding sample element $\fKrn_i(\vPaPrt_i)$:
\begin{align}
\label{eq:zero_bias}
\expect\left[ \fKrn_i(\vPaPrt_i) \vY_i \right] = \int \fKrn_i(\vPaPrt_i) \vY_i d\pP(\vPaPrt) = \vMatDiag_i.
\end{align}
If off-diagonal elements scale as $\sum_{j\neq i} \mMat_{ij}^2 = \vOmega_i^2 \vMatDiag_i^2$ for $i \in [\nPa]$,
then the zero-bias minimum-variance kernel is
\begin{align}
\label{eq:optimal_kernel}
\fKrn_i^*(\vPaPrt_i) = \frac{\krnCnst^{-1} \vPaPrt_i}{\vPaPrt_i^2 + \vOmega_i^2}
\quad\text{where}\quad
\krnCnst = \int \frac{\vPaPrt_i^2}{\vPaPrt_i^2 + \vOmega_i^2} d\pP(\vPaPrt_i) .
\end{align}
\end{theorem}

\newcommand{\prooftwo}{
\paragraph{Proof of \Cref{thm:optimal_kernel}}
We first note the elementwise expressions for matrix-vector products: $\vY_i = \sum_j \mMat_{ij} \vPaPrt_j$ and
\begin{align*}
\vY_i^2 = \sum_{j} [ \mMat_{ij}^2 \vPaPrt_j^2 + \sum_{k \neq j} \mMat_{ij} \mMat_{ik} \vPaPrt_j \vPaPrt_k ].
\end{align*}
As each density $\pP(\vPaPrt_j)$ has mean zero and unit variance,
we obtain the following integrals over all coordinates except $\vPaPrt_i$:
%Since each factor $\pP(\vPaPrt_j)$ has mean zero and unit variance,
%we obtain the following integrals over all coordinates except $\vPaPrt_i$:
\begin{align}
\label{eq:moments}
\int \left\{\vY_i,\: \vY_i^2 \right\} \prod_{j\neq i} d\pP(\vPaPrt_j) = \left\{\vMatDiag_i \vPaPrt_i,\: \vMatDiag_i^2 (\vPaPrt_i^2 + \vOmega_i^2) \right\}.
\end{align}
%\end{align}
Let $\fKrnPrt(\vPaPrt_i)$ be arbitrary feasible kernel variations near
$\fKrn_i^*(\vPaPrt_i)$ as $\fKrn_i(\vPaPrt_i) = \fKrn_i^*(\vPaPrt_i) + \eps \fKrnPrt(\vPaPrt_i)$,
where $\eps$ is a scalar differential element.
Differentiating w.r.t.~to $\eps$ yields G\^{a}teaux derivatives in any direction $\fKrnPrt(\vPaPrt_i)$.
To obtain an estimator with zero bias, all feasible kernels must satisfy \Cref{eq:zero_bias}.
Thus, 
\begin{align}
\label{eq:feasible_kernel_variations}
%\int \fKrn_i(\vPaPrt_i) \vY_i d\pP(\vPaPrt) &= \vMatDiag_i,
%\quad\text{thus}\\
\frac{\partial}{\partial \eps} \int \fKrn_i(\vPaPrt_i) \vY_i d\pP(\vPaPrt) &= \vMatDiag_i \int \fKrnPrt(\vPaPrt_i) \vPaPrt_i d\pP(\vPaPrt_i) = 0.
\end{align}
%\end{align}
Combining \Cref{eq:moments} with the estimator variance gives
\begin{align}
\variance &= \!\int\! \left[ \fKrn_i(\vPaPrt_i) \vY_i - \vMatDiag_{i} \right]^2 d\pP(\vPaPrt) = \!\int\! \fKrn_i^2(\vPaPrt_i) \vY_i^2 d\pP(\vPaPrt) - \vMatDiag_i^2 \nonumber\\
\label{eq:var_full}
&= \vMatDiag_i^2 \left[ \int \fKrn_i^2(\vPaPrt_i) (\vPaPrt_i^2 + \vOmega_i^2) d\pP(\vPaPrt_i) - 1 \right].
\end{align}
We can minimize $\variance$ by applying the variational principle
\begin{displaymath}
%\begin{align}
\frac{\partial \variance}{\partial \eps} = 2 \vMatDiag_i^2 \int \fKrnPrt(\vPaPrt_i) \fKrn_i^*(\vPaPrt_i) (\vPaPrt_i^2 + \vOmega_i^2) d\pP(\vPaPrt_i) = 0.
\end{displaymath}
%\end{align}
This must hold for all feasible $\fKrnPrt(\cdot)$ in \Cref{eq:feasible_kernel_variations},
thus $\fKrn_i^*(\vPaPrt_i) (\vPaPrt_i^2 + \vOmega_i^2) \propto \vPaPrt_i$.
Solving \Cref{eq:zero_bias} for the constant of proportionality $\krnCnst$ gives the stated result$\qed$
%thus \Cref{eq:feasible_full_pert} gives $\fKrn_i^*(\vPaPrt_i) (\vPaPrt_i^2 + \vOmega_i^2) \propto \vPaPrt_i \qed$
}
%\prooftwo

\Cref{thm:optimal_kernel} applies to any sample density with i.i.d.~factors that have zero mean and unit variance,
but it requires knowing the scaling factor $\vOmega_i^2$ for off-diagonal matrix elements.
Fortunately, the optimal distribution in \Cref{thm:optimal_density} obviates this issue.
Otherwise, it can be proven that $\omega_i^2 \leq 1$ for diagonally dominant matrices, providing a conservative choice.

See \appendixB~for a note regarding restricted updates, which further reduce estimation errors by hollowing out the update density.
Consistently, \Cref{thm:optimal_density} shows that the optimal density removes all small perturbations.

\begin{theorem}[Optimal Perturbation Density]
\label{thm:optimal_density}
From the set of perturbation densities that comprise a product of i.i.d.~factors,
$\pP(\vPaPrt) = \prod_{i=1}^\nPa \pP(\vPaPrt_i)$,
with zero mean and unit variance in each coordinate,
the optimal density to minimize the estimator variance in \Cref{thm:optimal_kernel}
is the Rademacher distribution in each factor,
i.e.~$\frac{1}{2}$ probability of either $-1$ or $1$.
Then, \Cref{eq:optimal_kernel} gives $\fKrn^*(\vPaPrt_i) = \vPaPrt_i$.
\end{theorem}

\newcommand{\proofthree}{
\paragraph{Proof of \Cref{thm:optimal_density}}
To apply the variational principle,
we write arbitrary density perturbations $\pPrt(\vPaPrt_i)$ in the vicinity of the optimizer $\pP^*(\vPaPrt_i)$
as $\pP(\vPaPrt_i) = \pP^*(\vPaPrt_i) + \eps \pPrt(\vPaPrt_i)$.
Since the variational principle need not hold for degrees of freedom on the boundary of the feasible set,
we will enforce the restriction $\pPrt(\vPaPrt_i) = 0$ when $\pP^*(\vPaPrt_i) = 0$.
If we only allow density perturbations that preserve normalization, we obtain 
\begin{align}
\label{eq:nrm_const}
\frac{\partial}{\partial \eps } \int d\pP(\vPaPrt_i) = \frac{\partial}{\partial \eps } 1 = \int d\pPrt(\vPaPrt_i) = 0.
\end{align}
Unpacking the estimator variance, \Cref{eq:var_full}, with the optimal kernel, \Cref{eq:optimal_kernel}, gives
\begin{displaymath}
\vMatDiag_i^{-2} \variance = \int \fKrn^{*2}(\vPaPrt_i) (\vPaPrt_i^2 + \vOmega_i^2) d\pP(\vPaPrt_i) - 1 = \krnCnst^{-1} - 1.
\end{displaymath}
As $\krnCnst$ is a function of $\pP(\cdot)$, differentiating w.r.t.~$\eps$ reveals the optimizer condition:
\begin{displaymath}
\vMatDiag_i^{-2} \frac{\partial \variance}{\partial \eps} = -\krnCnst^{-2} \int \frac{\vPaPrt_i^2}{\vPaPrt_i^2 + \vOmega_i^2} d\pPrt(\vPaPrt_i) = 0.
\end{displaymath}
This requires $\frac{\vPaPrt_i^2}{\vPaPrt_i^2 + \vOmega_i^2}$ to be a constant when $\pP^*(\vPaPrt_i) > 0$,
since \Cref{eq:nrm_const} provides the only constraint on feasible $\pPrt(\cdot)$ on this set.
Thus, only $\vPaPrt_i = \pm x$ for some $x \geq 0$ can map to nonzero density.
Solving for zero mean and unit variance gives the Rademacher distribution$\qed$ 
}
%\proofthree

Given optimal mechanisms to approximate the density of gradient variations,
we derive an upper bound on expected increases in loss.
The bound in \Cref{thm:glass_loss} follows from an assertion that predictive loss cannot become arbitrarily negative.
Rather, reductions due to small parameter changes are bound from below by a \textit{local floor},
the minimum loss that can be reached by a small change to a single parameter.

\begin{theorem}[Curvature of Expectation in Glass Loss]
\label{thm:glass_loss}
As we displace parameters by $\vPaPrt$,
the increase in loss $\dLoss(\vPaPrt) = \fLss(\vPaCnt + \vPaPrt) - \fLss(\vPaCnt)$ is greatest if a local floor enforces $\dLoss(\vPaPrt_k) \geq 0$.
Using the diagonal approximation of gradient variations, $\vGrdVar \approx \vGrdVarDns \oH |\vPaPrt|$ with $\vGrdVarDns \succcurlyeq 0$,
and treating corresponding loss effects independently in each coordinate,
we obtain the upper bound on expected increases in loss
\begin{align}
\label{eq:glass_loss}
%\expect\left[\dLoss(\vPaPrt)\right] \leq \sqrt{\frac{2}{3\pi} \vGrdVarDns^T |\vPaPrt|^3}.
\expect\left[\dLoss(\vPaPrt)\right] \leq \sqrt{\frac{2}{3\pi}} \vGrdVarDns^{\frac{1}{2}T} |\vPaPrt|^{\frac{3}{2}}.
\end{align}
\end{theorem}

\newcommand{\prooffour}{
\paragraph{Proof of \Cref{thm:glass_loss}}
To approximate the loss effects due to movement, $\vPa_k = \vPaCnt_k + \vPaPrt_k$,
we need to aggregate the effects of discrete gradient perturbations.
If we consider $n$ zero mean i.i.d.~$\vGrdPrt_k^{(j)}$ for $j = 1, \ldots, n$,
that activate at equally-spaced locations $\frac{j}{n} \vPaPrt_k$,
then we have $\expect[\vGrdPrt_k^2] = \frac{\vGrdVarDns_k |\vPaPrt_k|}{n}$
to preserve the density $\vGrdVarDns_k$.
Let $\dL_n(\vPaPrt_k)$ represent corresponding loss changes in the absence of a local floor.
Integration gives
\begin{displaymath}
\dL_n(\vPaPrt_k) = \sum_{j=1}^n \int_{\frac{j\vPaPrt_k}{n}}^{\vPaPrt_k} \vGrdPrt_k^{(j)} d\vPa_k 
= \sum_{j=1}^n \frac{n-j}{n} \vGrdPrt_k^{(j)} \vPaPrt_k.
\end{displaymath}
Since each term has zero mean, the variance is the sum. Thus
\begin{align*}
&\expect\left[ \dL_n(\vPaPrt_k)^2 \right] = \sum_{j=1}^n \frac{(n-j)^2}{n^2} \frac{\vGrdVarDns_k |\vPaPrt_k|}{n} \vPaPrt_k^2\quad\text{so} \\
&\expect\left[ \dL(\vPaPrt_k)^2 \right] = \lim_{n \to \infty} \frac{n - 3n^2 + 2n^3}{6 n^3} \vGrdVarDns_k |\vPaPrt_k|^3 = \frac{\vGrdVarDns_k |\vPaPrt_k|^3}{3}.
\end{align*}
This finite limit satisfies the Lyapunov formulation of the central limit theorem.
Thus, summing variance contributions gives $\dL(\vPaPrt_k) \sim \pN\left(0, \vGrdVarDns^T |\vPaPrt|^3/3 \right)$.

We enforce the local floor $\dLoss(\vPaPrt_k) \geq 0$ by mapping every virtual loss trajectory
that could be realized by $\dL(\vPaPrt_k)$
to a bounded trajectory $\dLoss(\vPaPrt_k) = \left|\dL(\vPaPrt_k)\right|$.
The stated result follows from $\int |x| d\pN(x \mid 0, \dist^2) = \sqrt{\frac{2}{\pi} \dist^2}\qed$  
}
%\prooffour

\Cref{thm:optimal_kernel,thm:optimal_density} are useful even when a glass term is not needed, since the resulting $\vHss$
accounts for distributed gradient changes that hold over a larger scale.
When present, the additional curvature due to \Cref{thm:glass_loss} reduces the length of an optimal step.
\Cref{thm:mod_qn} incorporates this additional curvature into QN steps that also include a non-negative $\vHss$.

\begin{theorem}[Optimal Modification to Quasi-Newton]
\label{thm:mod_qn}
By combining the bound in \Cref{eq:glass_loss} with the gradient $\vGrd$ and a nonnegative Hessian diagonal $\vHss \succcurlyeq 0$, so that
\begin{displaymath}
\fLss(\vPaCnt + \vPaPrt) \leq \fLss(\vPaCnt) + \vPaPrt^T \left(\vGrd + \frac{1}{2} \vHss \oH \vPaPrt \right) + \sqrt{\frac{2}{3\pi} \vGrdVarDns^T |\vPaPrt|^3},
\end{displaymath}
the minimizer is $\vPaPrt = -\vGrd \oH \vHssMod^{-1}$ using the modified Hessian
\begin{align}
\label{eq:mod_hess}
&\vHssMod = \vHssGls + \vHss + \sqrt{\vHssGls \oH (\vHssGls + 2 \vHss)} + \eps
\quad\text{where} \\
&\vHssGls = 3 \vGrdVarDns \oH \left(4 \pi |\fGrd(\vPaCnt)| + \eps\right)^{-1}.
\end{align}
We have added $\eps > 0$ terms for numerical stability.
\end{theorem}

\newcommand{\prooffive}{
\paragraph{Proof of \Cref{thm:mod_qn}}
We examine each step coordinate $\vPaPrt_k$ independently, since the effects are just added.
We can apply the variational principle by differentiating w.r.t.~$\vPaPrt_k$:
\begin{displaymath}
0 = \vGrd_k + \vHss_k \vPaPrt_k + \sqrt{\frac{3 \vGrdVarDns_k}{2\pi}} \sign(\vPaPrt_k) |\vPaPrt_k|^{1/2}.
\end{displaymath}
To enforce gradient descent, let $\vPaPrt_k = -\sign(\vGrd_k) x^2$ and take $x > 0$.
Substituting $\vHssGls_k$, we can obtain the expression
\begin{displaymath}
0 = -\sqrt{\frac{|\vGrd_k|}{2}} + \sqrt{\vHssGls_k} x + \frac{\vHss_k}{\sqrt{2|\vGrd_k|}} x^2.
\end{displaymath}
The quadratic formula, keeping $x > 0$, gives
\begin{displaymath}
x = \frac{-\sqrt{\vHssGls_k} + \sqrt{\vHssGls_k + 2 \vHss_k}}{\vHss_k \sqrt{\frac{2}{|\vGrd_k|}}}
=\frac{\sqrt{2|\vGrd_k|}}{\sqrt{\vHssGls_k} + \sqrt{\vHssGls_k + 2 \vHss_k}}
\end{displaymath}
Substitution into $\vPaPrt_k$ yields the stated result$\qed$. 
}
%\prooffive

% ===================================================
% ======== Nesterov Accelerated Quasi-Newton ========
% ===================================================
Unfortunately, a simple example shows that QN steps are often too greedy, especially for over-parameterized models:
take an underdetermined linear system comprising a random matrix $\mA \in \reals^{10 \times 100}$ and vector $\vB \in \reals^{10 \times 1}$.
For $\vX \in \reals^{100}$, let the loss be $\fLss(\vX) = \frac{1}{2} \left\| \mA \vX - \vB \right\|_2^2$
with gradient $\fGrd(\vX) = \mA^T (\mA \vX - \vB)$ and Hessian diagonal $\vHss = \diag(\mA^T \mA)$.
In an actual (first) trial, a QN step from zero gave $\|\vX\|_2 = 3.75$ with an \textit{increase} in loss from $5.6$ to $464.6$.
The minimal solution needed only $\|\vX_{\text{min}}\|_2 = 1.1$,
showing off-diagonal terms' importance.
Fortunately, damped QN steps solve this problem; only one $10\%$ step reduced the loss to $0.8$.
%there are other ways to solve this problem.
%For example, damping QN down to $10\%$ reduced the loss to $0.8$ in one step.

\paragraph{Nesterov Accelerated Quasi-Newton}
Damped QN steps create an opportunity to not only reduce the loss more efficiently, but also improve the correctness of gradient updates with NAQ.
The approaches taken by \Citet{ninomiya2017novel} and \citet{indrapriyadarsini2020stochastic} scale QN steps with an Armijo line search,
but we show how a simple relationship with the gradient momentum coefficient, $\beta_1$ in Adam \citep{Kingma2014}, provides desirable exactness properties.
To understand this, we briefly review some related optimization practices.
See \citet{ruder2016overview} for an overview.

Nesterov acceleration anticipates gradient changes by distinguishing the actual state of an optimization trajectory from the location of gradient evaluations.
In our notation, at step $s$ we have a tentative parameter update $\vPaPrt^{(s)}$.
The actual parameter position $\vPaCnt^{(s)}$ will be updated by taking a fraction $\varphi \leq 1$ of the full step,
but the next gradient will be evaluated at a new location $\vPaEvl^{(s+1)}$ with a different factor $\omega \geq \varphi$:
\begin{align}
\label{eq:pa_updates}
\vPaCnt^{(s+1)} \!= \vPaCnt^{(s)} \!+ \varphi \vPaPrt^{(s)}
\quad\text{vs}\quad
\vPaEvl^{(s+1)} \!= \vPaCnt^{(s)} \!+ \omega \vPaPrt^{(s)}.
\end{align}
These updates combine with gradient momentum as
\begin{align}
\label{eq:grad_update}
\vGrd^{(s+1)} = \beta_1 \vGrd^{(s)} + (1 - \beta_1) \fGrd(\vPaEvl^{(s)}).
\end{align}

Going forward, we use the modified Hessian $\vHssMod$ from \Cref{eq:mod_hess} with the quadratic loss approximation
\begin{align}
\label{eq:quad_loss}
\fLss(\vPaCnt^{(s)} + \vPaPrt) = \fLss(\vPaCnt^{(s)}) + \vPaPrt^T \left(\vGrd^{(s)} + \vHssMod \oH \vPaPrt / 2 \right),
\end{align}
which matches the gradient at $\vPaCnt$ while keeping the optimal step.
With this setup, \Cref{thm:naq} shows how certain $\varphi$ and $\omega$ yield exactness properties in \Cref{eq:grad_update}.

%lookahead strategies \citep{zhang2019lookahead} 
%\citet{rumelhart1986learning} introduced a momentum term in weight updates that is (equivalent?) to the damped average gradient.  

\begin{theorem}[Exact Nesterov Accelerated Quasi-Newton]
\label{thm:naq}
For the purpose of optimization, we apply the loss approximation in \Cref{eq:quad_loss} with the matching gradient
\begin{align}
\label{eq:quad_loss_grad}
\fGrd(\vPaCnt^{(s)} + \vPaPrt) = \vGrd^{(s)} + \vHssMod \oH \vPaPrt,
\end{align}
and optimal step, $\vPaPrt^{(s)} = -\vGrd^{(s)} \oH \vHssMod^{-1}$.
Yet, suppose the true gradient has unknown linear dependence
\begin{align}
\label{eq:hidden_grad}
\fGrd^*(\vPaCnt^{(s)} + \vPaPrt) = \fGrd^*(\vPaCnt^{(s)}) + \mHss \vPaPrt
\end{align}
that holds through displacements to $\vPaPrt^{(s)}$, our projected optimum.
Then, the Nesterov acceleration coefficients,
\begin{align}
\label{eq:optimal_naq}
\varphi = 1 - \beta_1
\quad\text{and}\quad
\omega = 1,
\end{align}
ensure that the running gradient reduction by $\beta_1$ in \Cref{eq:grad_update} matches \Cref{eq:quad_loss_grad}
and the new gradient evaluation exactly captures the unknown linear dependence in \Cref{eq:hidden_grad},
while suppressing errors $\vGrdPrt^{(s)}$ so that 
%(both off-diagonal and diagonal discrepancies)
%to suppress running errors so that
\begin{align}
\label{eq:error_reduction}
\vGrd^{(s)} &= \fGrd^*(\vPaCnt^{(s)}) + \vGrdPrt^{(s)}
\quad\text{goes to} \nonumber\\
\vGrd^{(s+1)} &= \fGrd^*(\vPaCnt^{(s+1)}) + \beta_1 \vGrdPrt^{(s)}.
\end{align}
\end{theorem}

\newcommand{\proofsix}{
\paragraph{Proof of \Cref{thm:naq}}
The first claim easily follows by unpacking \Cref{eq:pa_updates} into \Cref{eq:quad_loss_grad}:
\begin{displaymath}
\fGrd(\vPaCnt^{(s+1)}) = \vGrd^{(s)} + \varphi \vHssMod \oH \vPaPrt^{(s)} = (1 - \varphi) \vGrd^{(s)} = \beta_1 \vGrd^{(s)}.
\end{displaymath}
Substituting the new gradient evaluation into \Cref{eq:hidden_grad},
using the error expression at step $s$ from \Cref{eq:error_reduction},
and placing the result in \Cref{eq:grad_update} gives
\begin{align*}
&\vGrd^{(s+1)}
= \beta_1 \vGrd^{(s)} + (1 - \beta_1) \left( \vGrd^{(s)} - \vGrdPrt + \omega \mHss \vPaPrt^{(s)} \right) \\
&= \left[\vGrd^{(s)} - \vGrdPrt \right] + \varphi \mHss \vPaPrt^{(s)} + \beta_1 \vGrdPrt
= \vGrd^*(\vPaCnt^{(s+1)}) + \beta_1 \vGrdPrt\qed
\end{align*}
}
%\proofsix

\Cref{thm:naq} provides updates that capture hidden linear discrepancies (diagonal and off-diagonal) averaged over the distance to the optimum we anticipate from our current knowledge.
Since the gradient damping is consistent with our loss topography estimate,
the following methods extinguish gradient information at the same rate it is exploited.
%Note that the association between $\beta_1$ and $\varphi$ may increase the importance of a hyperparameter search on $\beta_1$.

% ==============================================================
% =========================== Methods ==========================
% ==============================================================
\section{Methods and Experiments}
\label{sec:methods}

Alice allows us to study which curvature computations are most suitable for capturing an architecture's loss topography,
with follow-on implications for sensitivity-based pruning and quantization.
By showing the impact that these curvature computations have on early training,
Alice identifies what is needed for longer or more complex procedures.

To compute three parameter-length quantities $\vGrd$, $\vGrdVarDns$, and $\vHss$,
we need three gradient evaluations.
To be consistent with \Cref{eq:pa_updates}, these are anchored to $\vPaEvl$ rather than $\vPaCnt$:
\begin{align}
\label{eq:grad_eval_plus}
\vGrd^{(+)} &= \fGrd(\vPaEvl + \dist \vPaPrt) = \fGrd(\vPaEvl) + \dist \mHss \vPaPrt + \vGrdPrt^{(+)} \\
\label{eq:grad_eval_minus}
\vGrd^{(-)} &= \fGrd(\vPaEvl - \dist \vPaPrt) = \fGrd(\vPaEvl) - \dist \mHss \vPaPrt + \vGrdPrt^{(-)},\quad\text{and} \\
\label{eq:grad_eval_zero}
\vGrd^{(0)} &= \fGrd(\vPaEvl).
\end{align}
These expressions capture average linear dependencies as matrix-vector multiplies, $\mHss \vPaPrt$.
Similarly, the gradient perturbations $\vGrdPrt^{(\pm)}$ reveal glass density $\expect[\vGrdPrt^{(\pm)2}] = \mGrdVarDns |\vPaPrt|$:
\begin{align}
\label{eq:hess_matvec}
\mHss \vPaPrt &= \expect\left[ \left(\vGrd^{(+)} - \vGrd^{(-)}\right)/(2\dist) \right],\quad\text{and}\\
\label{eq:GVD_matvec}
\mGrdVarDns |\vPaPrt| &= \expect\left[ \left( \left(\vGrd^{(+)} + \vGrd^{(-)}\right)/2 - \vGrd^{(0)}\right)^2/(2\dist) \right].
\end{align}
We can apply \Cref{thm:optimal_density} without storing $\vPaPrt$ by enforcing non-negative $\vHss$,
since $\left|\vPaPrt \oH \mHss \vPaPrt\right| = |\mHss \vPaPrt|$,
which also repels saddle points.
\Cref{alg:topo_update} shows how we incorporate these evaluations into running topography updates.

Alice can also enable quick-steps that avoid updating the Hessian and glass densities too often.
For example, taking 3 regular steps (1 gradient each) between every full step (3 gradients)
reduces costs to 6 gradients every 4 steps (i.e.~half of full extraction on every step).
The optimization impact can be small in comparison to speed improvements.

\begin{algorithm}[t]
\caption{Alice Topography Update}
\label{alg:topo_update}
\begin{minipage}{0.98\columnwidth}
\textbf{Input}:
evaluation center: $\vPaEvl$;
gradient function: $\fGrd(\vPa)$.\\
\textbf{Input and Output}: running averages: $\vGrd$, $\vGrdVarDns$, and $\vHss$.\\
\textbf{Hyperparameters}: $\dist, \beta_1, \beta_2$.
\end{minipage}
\begin{algorithmic}[1] %[1] enables line numbers
\STATE Draw Rademacher $\vT$.
\STATE Evaluate $\vGrd^{(\pm)} = \fGrd(\vPaEvl \pm \dist \vT)$ and $\vGrd^{(0)} = \fGrd(\vPa)$.
\STATE $\vGrd \gets \beta_1 \vGrd + (1 - \beta_1) \vGrd^{(0)}$. 
\STATE $\vHss \gets \beta_2 \vHss + (1 - \beta_2) \frac{1}{2 \dist}\left| \vGrd^{(+)} - \vGrd^{(-)} \right|$. \label{line:hess}
\STATE $\vGrdVarDns \gets \beta_2 \vGrdVarDns + (1 - \beta_2) \frac{2}{\dist} \left(\frac{1}{2} (\vGrd^{(+)} + \vGrd^{(-)}) - \vGrd^{(0)}\right)^2$. \label{line:glass}
\end{algorithmic}
\end{algorithm}

\begin{figure}[t]
\centering
\includegraphics[width=0.98\columnwidth]{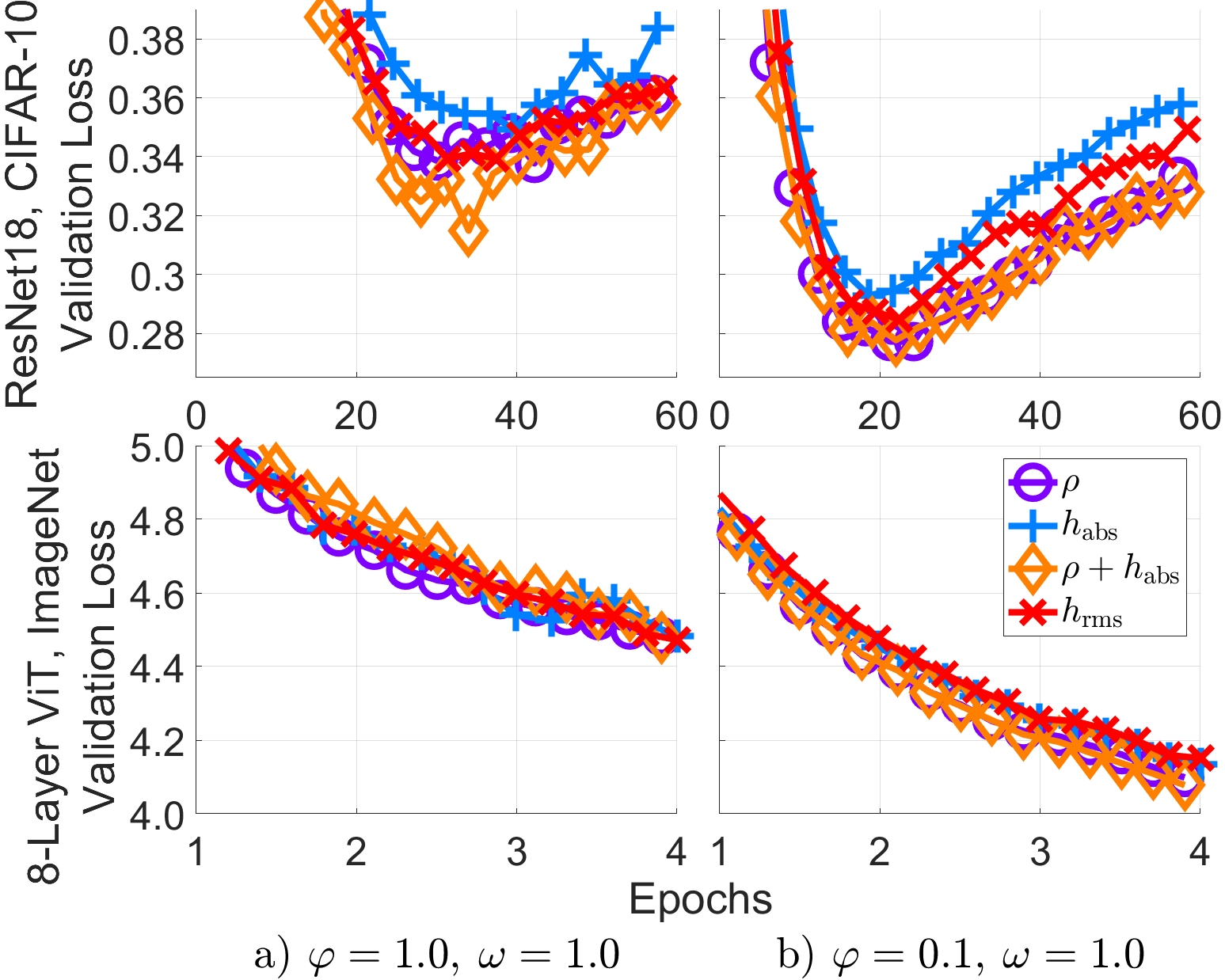} 
\caption{Alice training with various curvature terms:
a) full QN steps, b) NAQ steps (\Cref{thm:naq}).}
\label{fig:naq}
\end{figure}

\paragraph{Nesterov Acceleration}
\Cref{fig:naq} examines the efficacy of NAQ with various curvature terms:
$\vHss_{\rm abs}$ is $|\vH|$ (\Cref{alg:topo_update}, Line \ref{line:hess})
and $\vRho$ indicates glass density (Line \ref{line:glass}).
We also include an RMS version $\vHss_{\rm rms}$ that is similar to AdaHessian \citep{yao2021adahessian}.
Our ViT model uses a linear-complexity form of attention \citep{shen2021efficient}
trained with the $64\times64$ downsampled Imagenet \citep{deng2009imagenet,chrabaszcz2017downsampled}.
The first column shows full QN steps,
and the second employs the NAQ coefficients in \Cref{eq:optimal_naq}.
Both NAQ and curvature using glass terms yield the best results.
%Perhaps the reason glass curvature alone yields good results may be due to NAQ capturing hidden linear terms missed by the glass.
%We speculate that the reason glass curvature alone can be sufficient to obtain good results
%may be due to NAQ capturing hidden linear terms missed by the glass.

\begin{figure}[t]
\centering
\includegraphics[width=0.98\columnwidth]{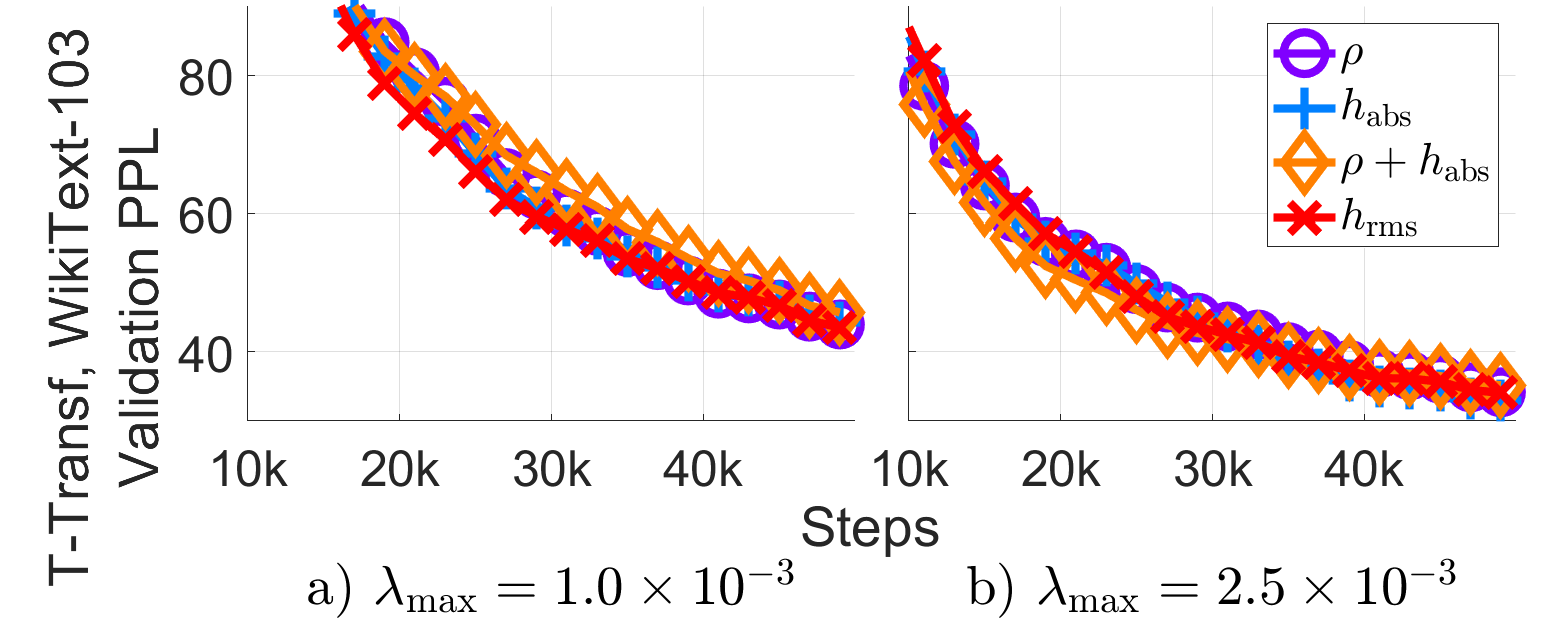} 
\caption{Training effect of increased exploitation bound.}
\label{fig:es}
\end{figure}

\paragraph{Exploration and Stability}
In principle, tracking loss curvature does not preclude traps (e.g., local minima or flat saddle points) from blocking progress.
Moreover, \Cref{thm:mod_qn}'s optimal step can become very large and exceed trustworthy extrapolation.
Alice addresses these problems using learning rate bounds $\lrMin$ and $\lrMax$
that can be interpreted as: 1.~fixed step lengths, 2.~SGD-M learning rates, or 3.~Adam learning rates.
Between these bounds, Alice takes NAQ steps.
This simple approach ensures that updates exploiting curvature remain sufficiently large and stable.
%A key property of Adam is how (regardless of local topography) the step factor
%$|\vPaPrt| =  \dist |\vGrd| \oH (\sqrt{\vSec} + \eps)^{-1}$ ensures $\expect\left[ \vPaPrt^2 \right] = \lambda^2$.
%Alice extends this functionality by disentangling exploration and stability
%with bounding factors, $\lrMin$ and $\lrMax$, that multiply the typical Adam step.

\Cref{fig:es} shows the importance of stability limits for early loss reduction using the Tensorised Transformer \citep{ma2019tensorized} with the WikiText-103 corpus \citep{merity2016pointer} and Adam-based bounds.
Here, increasing $\lrMax$ offers a clear benefit. 
%Although learning-rate schedules can eventually recover near-optimal outcomes, the right step bounds can accelerate early improvements.
Conveniently, by setting $\varphi = \omega = 1$ and $\lrMin = \lrMax = $ learn-rate, Alice replicates either SGD-M or Adam.
Thus, Alice can always perform at least as well as these methods,
and further adjustments offer a possibility of improvement.

\begin{figure}[t]
\centering
\includegraphics[width=0.98\columnwidth]{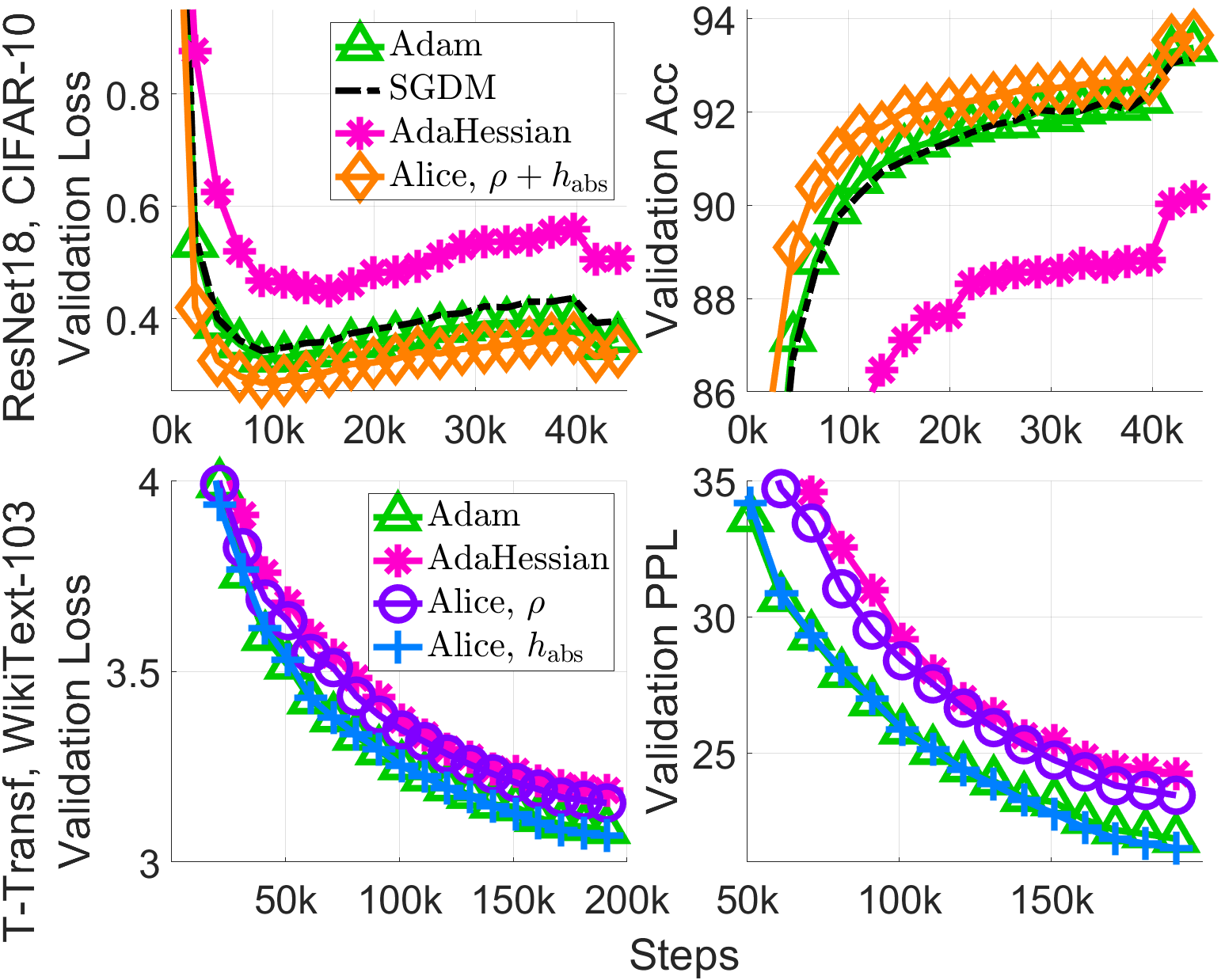} 
\caption{Optimization method comparisions.}
\label{fig:rn_methods}
\end{figure}

\paragraph{Method Comparisons}
Although our aim is to lay groundwork for additional optimization challenges, 
we include some comparisons against related optimizers for basic loss reduction with dense parameters.
\Cref{fig:rn_methods} shows validation quality on ResNet18 trained with stochastic gradient descent with momentum (SGD-M),
Adam, Adahessian, and Alice.

% ==================================================================
% =========================== Experiments ==========================
% ==================================================================
%\section{Experiments}
%\label{sec:experiments}

%\paragraph{Exploration and Stability}
%Here we examine 

% =================================================================
% =========================== Discussion ==========================
% =================================================================
\section{Summary}
\label{sec:discussion}

\paragraph{Summary}
We developed a mathematical framework to understand the influence of dense gradient discontinuities on the loss topography.
These discontinuities are caused by ReLU-like activation functions,
which form a gradient glass when they occur in high numbers.
We then derived the optimal kernel and sample density for estimating locally-averaged diagonal dependencies from matrix-vector products,
allowing us to approximate linear gradient dependence (i.e., the locally-averaged Hessian diagonal)
and gradient variations (i.e., the local glass density) from ordinary backpropagation.
We also showed how expected changes in the loss are bound by a $\frac{3}{2}$ power law.

By including glass terms in the loss, we derived the optimal modification to quasi-Newton steps
to unify model updates in architectures that may have gradient effects that are dominated by either term.
Further, our analysis of Nesterov acceleration finds that certain coefficients can exactly capture the gradient reductions we expect,
while also correcting hidden linear terms in updates to a running gradient average.

Alice provides a baseline to show how these techniques can improve basic loss reduction.
By examining the efficacy of different curvature approximations,
we gain insight into the loss landscape that will support advanced optimization and regularization strategies currently being explored.

% ============================
% ======== Conclusion ========
% ============================
%\subsection{Conclusion}

%Lorem ipsum treguna mekoides trecorum satis dee finite incantatem.

% ==========================================================
% ======================== Appendix ========================
% ==========================================================
\appendix

% ========================================================
% ======================== Proofs ========================
% ========================================================
\section{Proofs}
\label{sec:proofs}

\proofone

\prooftwo

\proofthree

\prooffour

\prooffive

\proofsix

\newcommand{\extended}{
\section{Methodological Details}

\begin{algorithm}[ht]
\caption{Alice Topography Update, Full Version}
\label{alg:topo_update_full}
\begin{minipage}{0.98\columnwidth}
\textbf{Input}:
evaluation center: $\vPaEvl$;
gradient function: $\fGrd(\vPa)$.\\
\textbf{Input and Output}: Averages: $\vGrd$, $\vGrdVarDns$, $\vHss_{\rm abs}$,$\vHss_{\rm rms}^2$, and $\vSec$.\\
\textbf{Hyperparameters}: $\dist, \beta_1, \beta_2$.\\
This construction uses \Cref{eq:grad_eval_plus,eq:grad_eval_minus,eq:grad_eval_zero,eq:hess_matvec,eq:GVD_matvec}
to adapt \Cref{thm:relu_pert,thm:optimal_kernel,thm:optimal_density} with only one parameter-length temporary $\vT$.
\end{minipage}
\begin{algorithmic}[1] %[1] enables line numbers
\STATE Draw Rademacher $\vT$. Set parameters $\vPa \gets \vPaEvl + \dist \vT$.
\STATE Evaluate $\fGrd(\vPa) = \vGrd^{(+)}$.
\STATE Set parameters $\vPa \gets \vPaEvl - \dist \vT$. Store $\vT \gets \vGrd^{(+)}$.
\STATE Evaluate $\fGrd(\vPa) = \vGrd^{(-)}$.
\STATE Update $\vHss_{\rm abs} \gets \beta_2 \vHss_{\rm abs} + (1 - \beta_2) \frac{1}{2 \dist}\left| \vT - \vGrd^{(-)} \right|$.
\STATE Update $\vHss_{\rm rms}^2 \gets \beta_2 \vHss_{\rm rms}^2 + (1 - \beta_2) \frac{1}{4 \dist^2}\left( \vT - \vGrd^{(-)} \right)^2$. \label{line:rms}
\STATE Store $\vT \gets \frac{1}{2} (\vT + \vGrd^{(-)})$. Set parameters $\vPa \gets \vPaCnt$.
\STATE Evaluate $\fGrd(\vPa) = \vGrd^{(0)}$.
\STATE Update $\vGrd \gets \beta_1 \vGrd + (1 - \beta_1) \vGrd^{(0)}$. 
\STATE Update $\vGrdVarDns \gets \beta_2 \vGrdVarDns + (1 - \beta_2) \frac{2}{\dist} \left(\vT - \vGrd^{(0)}\right)^2$.
\STATE Update $\vSec \gets \beta_2 \vSec + (1 - \beta_2) \vGrd^{(0)2}$. \label{line:second}
\end{algorithmic}
\end{algorithm}

\begin{algorithm}[ht]
\caption{Alice Optimization Step}
\label{alg:step}
\begin{minipage}{0.98\columnwidth}
\textbf{Input}: running averages: $\vGrd$, $\vGrdVarDns$, $\vHss$, and $\vSec$.\\
\textbf{Input and Output}: parameters: $\vPaCnt$.\\
\textbf{Output}: evaluation center: $\vPaEvl$.\\
\textbf{Hyperparameters}: limit-method, $\dist, \lrMin, \lrMax, \eps$.\\
These parameter updates enable \Cref{thm:glass_loss,thm:mod_qn,thm:naq} while controlling exploitation stability with Adam-based limits.
\end{minipage}
\begin{algorithmic}[1] %[1] enables line numbers
\STATE Compute glass term, $\vHssGls = \frac{3}{4\pi}\vGrdVarDns \oH \left(|\vGrd| + \eps \right)^{-1}$, and
\STATE modified Hessian, $\vHssMod = \vHssGls + \vHss + \sqrt{\vHssGls \oH (\vHssGls + 2 \vHss)} + \eps$.
\STATE Get quasi-Newton scale, $\vPaPrt = |\vGrd| \oH \vHssMod^{-1}$, and
\STATE Fixed scale limits, $\vPaPrt_{\rm min/max} = \dist_{\rm min/max}$; or
\STATE SGD-M limits, $\vPaPrt_{\rm min/max} = \dist_{\rm min/max} |\vGrd|$; or
\STATE Adam limits, $\vPaPrt_{\rm min/max} = \dist_{\rm min/max} |\vGrd| \oH \left( \sqrt{\vSec} + \eps \right)^{-1}$.
\STATE Enforce bounds, $\vPaPrt \gets \max\left(\vPaPrt_{\rm min}, \min\left( \vPaPrt_{\rm max}, \vPaPrt\right)\right)$.
\STATE Correct the sign for descent, $\vPaPrt \gets -\sign(\vGrd) \vPaPrt$.
\STATE Update evaluation center, $\vPaEvl \gets \vPaCnt + \omega \vPaPrt$.
\STATE Update parameters, $\vPaCnt \gets \vPaCnt + \varphi \vPaPrt$.
\end{algorithmic}
\end{algorithm}

\paragraph{Restricted Updates:}
When a perturbation contains a relatively small value of $|\vPaPrt_i|$,
the corresponding element of the matrix-vector product $\vY_i$ expresses only a small contribution from the diagonal $\vMatDiag_i$.
With this in mind, we observe the variance of the diagonal estimator can be further reduced
by rejecting updates to a running $\vMatDiag_i$ estimate from samples with $|\vPaPrt_i|$ below a threshold $\dropThres$.
The reason an improvement is possible is because these rejections change the active density from $\pP(\vPaPrt_i)$ to a restricted density
\begin{displaymath}
%\label{eq:restricted_distribution}
\pR(\vPaPrt_i) \propto
\begin{cases}
\pP(\vPaPrt_i) & |\vPaPrt_i| \geq \dropThres \\
0        & \text{otherwise}.
\end{cases}
\end{displaymath}
%\end{align}

For example, if we average $\nSa$ standard normal samples
%, $\pP(\vPaPrt) \equiv \pN(\vPaPrt \mid 0, \mI)$,
and set $\omega_i^2 = 1$, then \Cref{thm:optimal_kernel} gives 
\begin{displaymath}
%\begin{align}
\fKrn_i^*(\vPaPrt_i) = \frac{2 \vPaPrt_i}{\vPaPrt_i^2 + 1}
\quad\text{with variance}\quad
\variance = \frac{3}{\nSa}\vMatDiag_i^2.
\end{displaymath}
%\end{align}
Here, we have divided the single-sample variance by the number of samples.
By sampling from the same distribution, but only updating if $|\vPaPrt_i| \geq 1$,
the optimal kernel becomes:
\begin{displaymath}
%\begin{align}
\fKrn_i^*(\vPaPrt_i) = \frac{1.40 \vPaPrt_i}{\vPaPrt_i^2 + 1}
\quad\text{with}\quad
\variance = \frac{0.823}{0.317\nSa}\vMatDiag_i^2 = \frac{2.60}{\nSa}\vMatDiag_i^2.
\end{displaymath}
%\end{align}
Note the reduction in effective samples to account for the update probability: $31.7\%$.
Nevertheless, these restricted updates improve the combined result.

\paragraph{Algorithms:}
\Cref{alg:topo_update_full} shows the full topography update computations used by Alice in a formulation that only requires one additional parameter-length temporary variable.
Alice also enables an RMS Hessian approximation shown in \Cref{alg:topo_update_full}, Line \ref{line:rms}.
%\begin{displaymath}
%[\vHssRms^2] \gets \beta_2 [\vHssRms^2] + (1 - \beta_2) \frac{1}{4 \dist^2}\left(\vT - \vGrd^{(-)}\right)^2.
%\end{displaymath}
Taking the square root of the running second moment yields a non-negative approximation that is suitable for QN steps.
This approach is similar to that of AdaHessian \citep{yao2021adahessian}, except they average results over parameter blocks.

\Cref{alg:step} shows the modified quasi-Newton steps for both glass density $\vGrdVarDns$ and a Hessian diagonal $\vHss$,
but either term can be set to zero in practice.
Then, the Adam factor and step bounds are applied before correcting the sign for gradient descent.

\section{Experimental Setup}

Our experiments leverage existing benchmarks and we report the minimum, median, and maximum relevant outcomes for each case.
Not only does this express the full range of outcomes, it also indicate the best quality model that would be used in practice. 

\paragraph{Hardware and Software}
Tests are performed on compute nodes that each contain AMD 32-core CPUs and 1 NVidia A100 GPU with 40GB of memory.
Software is written for PyTorch 2.2.1 with Cuda 11.8 drivers.  

\paragraph{ResNet18, Power Law:}
The results in \Cref{fig:grad_var_exp} are generated with a specialized version of Alice that uses extra memory and gradient evaluations to compute and store gradient variations at $\dist$ and $2\dist$.
Training is performed on a modified version of ResNet18 with multiplicative masks on channels, which improves prediction quality for SGDM, Adam, and Alice.
We train for 40 epochs using $\dist = 0.002$, $\beta_1 = 0.9$, $\beta_2 = 0.999$, $\eps = 10^{-8}$, $\varphi = 0.1$, $\omega = 1.0$, $\lrMin = 0$, $\lrMax = 0.002$, ${\rm quick\_steps} = 0$, and Adam-based limiting.
Step curvature uses both $\vGrdVarDns$ and $\vHss_{\rm abs}$.

\paragraph{ResNet18, NAQ:}
The ResNet18 trials in \Cref{fig:naq} use the same settings as above, except with the tuned learning rate of $\dist = 0.005$, fixed-scale limiting with $\lrMax=0.005$, and trained for 60 epochs. These options were tuned from the set $\{0.001, 0.002, 0.005, 0.01, 0.02, 0.05\}$ and trying both Adam-based limiting and fixed scale limiting with $\lrMax=\dist$ or $\lrMax=2\dist$.
Results include 10 seeds.

\begin{center}
\begin{tabular}{|c c l | c c c |}
\multicolumn{3}{c}{Settings} & \multicolumn{3}{c}{Test Accuracy} \\ 
\hline
$\varphi$ & $\omega$ & Terms & Min & Median & Max \\ \hline
1.0 & 1.0 &               $\rho$ &  90.6\% &  91.2\% &  91.5\% \\ \hline
1.0 & 1.0 &        $h_{\rm abs}$ &  89.7\% &  90.1\% &  90.8\% \\ \hline
1.0 & 1.0 & $\rho + h_{\rm abs}$ &  91.0\% &  91.8\% &  92.7\% \\ \hline
1.0 & 1.0 &        $h_{\rm rms}$ &  90.5\% &  90.8\% &  91.2\% \\ \hline
0.1 & 1.0 &               $\rho$ &  92.1\% &  \textbf{92.6\%} &  92.7\% \\ \hline
0.1 & 1.0 &        $h_{\rm abs}$ &  91.8\% &  91.9\% &  92.1\% \\ \hline
0.1 & 1.0 & $\rho + h_{\rm abs}$ &  92.1\% &  92.4\% &  \textbf{92.8\%} \\ \hline
0.1 & 1.0 &        $h_{\rm rms}$ &  91.8\% &  92.3\% &  92.7\% \\ \hline
\end{tabular}
\end{center}

\paragraph{ViT, NAQ:}
Our vision transformer trials in \Cref{fig:naq} use 8 encoding layers based on a linear-complexity attention formulation \citep{shen2021efficient}
and trained with the $64\times64$ downsampled Imagenet \citep{deng2009imagenet,chrabaszcz2017downsampled}.
We train for 4 epochs using $\dist = 0.005$, $\beta_1 = 0.9$, $\beta_2 = 0.999$, $\eps=10^{-8}$, $\lrMin = 0$, $\lrMax = 0.01$, ${\rm quick\_steps} = 3$,
and Adam-based limiting.
Results include 5 seeds.

\begin{center}
\begin{tabular}{|c c l | c c c |}
\multicolumn{3}{c}{Settings} & \multicolumn{3}{c}{Test Loss} \\  
\hline
$\varphi$ & $\omega$ & Terms & Min & Median & Max \\ \hline
1.0 & 1.0 &               $\rho$ &  4.49 &  4.49 &  4.55 \\ \hline
1.0 & 1.0 &        $h_{\rm abs}$ &  4.45 &  4.49 &  4.73 \\ \hline
1.0 & 1.0 & $\rho + h_{\rm abs}$ &  4.41 &  4.48 &  4.48 \\ \hline
1.0 & 1.0 &        $h_{\rm rms}$ &  4.46 &  4.47 &  4.65 \\ \hline
0.1 & 1.0 &               $\rho$ &  4.08 &  4.09 &  4.12 \\ \hline
0.1 & 1.0 &        $h_{\rm abs}$ &  4.13 &  4.13 &  4.14 \\ \hline
0.1 & 1.0 & $\rho + h_{\rm abs}$ &  \textbf{4.05} &  \textbf{4.07} &  4.11 \\ \hline
0.1 & 1.0 &        $h_{\rm rms}$ &  4.14 &  4.15 &  4.15 \\ \hline
\end{tabular}
\end{center}

\paragraph{Tensorized Transformer, Stability Limit:}
Our Tensorised Transformer trials use the architecture described by \citet{ma2019tensorized}, with a dropout rate of 0.1 and batch-size 60 for 50,000 steps.
The results shown in \Cref{fig:es} are obtained from Alice with $\dist = 0.001$, $\beta_1 = 0.9$, $\beta_2 = 0.999$, $\eps = 10^{-8}$, $\varphi = 0.1$, $\omega = 1.0$, ${\rm quick\_steps} = 3$, and Adam-based limiting.
Results include 10 seeds.
It remains unclear why NAQ is not as effective as full quasi-Newton steps for this problem.

\begin{center}
\begin{tabular}{|c l | c c c |}
\multicolumn{2}{c}{Settings} & \multicolumn{3}{c}{Validation PPL} \\  
\hline
$\lrMax$ & Terms & Min & Median & Max \\ \hline
$1.0\times 10^{-3}$ &               $\rho$ &  40.4 &  43.2 &  49.1 \\ \hline
$1.0\times 10^{-3}$ &        $h_{\rm abs}$ &  38.6 &  43.7 &  47.1 \\ \hline
$1.0\times 10^{-3}$ & $\rho + h_{\rm abs}$ &  39.8 &  45.1 &  48.2 \\ \hline
$1.0\times 10^{-3}$ &        $h_{\rm rms}$ &  39.4 &  42.9 &  51.2 \\ \hline
$2.5\times 10^{-3}$ &               $\rho$ &  \textbf{28.8} &  33.6 &  35.8 \\ \hline
$2.5\times 10^{-3}$ &        $h_{\rm abs}$ &  30.0 &  \textbf{32.5} &  36.8 \\ \hline
$2.5\times 10^{-3}$ & $\rho + h_{\rm abs}$ &  31.0 &  34.8 &  39.1 \\ \hline
$2.5\times 10^{-3}$ &        $h_{\rm rms}$ &  28.9 &  33.7 &  39.9 \\ \hline
\end{tabular}
\end{center}

\paragraph{ResNet18, Training Methods:}
For ResNet18 results in \Cref{fig:rn_methods}, the Adam trials use $\dist = 0.001$, $\beta_1 = 0.9$, $\beta_2 = 0.999$, and $\eps=1E^{-8}$.
SGD-M uses $\dist = 0.001$ and $\beta = 0.9$.
AdaHessian uses $\dist = 0.15$, which gave the best results from the set $\dist \in \{ 0.1, 0.15, 0.2, 0.25, 0.3 \}$.
Alice uses $\dist = 5e-3$, $\beta_1 = 0.9$, $\beta_2 = 0.999$, $\eps=1E^{-8}$, $\varphi = 0.1$, $\omega = 1.0$, $\lrMin = 0$, $\lrMax = 0.01$, ${\rm quick\_steps} = 3$, and Adam-based limiting.
Curvature includes both $\vGrdVarDns$ and $\vHss_{\rm abs}$.
Results include 30 seeds.

\begin{center}
\begin{tabular}{| c |c c c|}
\multicolumn{1}{c}{Method} & \multicolumn{3}{c}{Test Accuracy} \\  
\hline
                               & Min & Median & Max \\ \hline
                          Adam & 93.23\% & 93.41\% & 93.73\% \\ \hline
                          SGDM & 92.65\% & 93.19\% & 93.75\% \\ \hline
                    AdaHessian & 78.41\% & 90.25\% & 91.50\% \\ \hline
   Alice, $\rho + h_{\rm abs}$ & 93.50\% & \textbf{93.76\%} & \textbf{93.96\%} \\ \hline
\end{tabular}
\end{center}

\paragraph{Tensorised Transformer, Training Methods:}
The Tensorised Transformer results in \Cref{fig:rn_methods} use the same settings as above, but taking 200,000 steps.
Adam and Alice use $\dist = 0.00025$, $\beta_1 = 0.9$, $\beta_2 = 0.999$, and $\eps=1E^{-8}$,
but Alice also includes $\varphi = 1.0$, $\omega = 1.0$, $\lrMax = 0.000375$, ${\rm quick\_steps} = 1$, and Adam-based limiting.
AdaHessian uses $\dist = 0.00025$.
Results include 5 seeds.

%\begin{center}
%\begin{tabular}{| c |c c c|} 
%\hline
%Test Loss & Alice  & Adam   & AdaHessian \\ \hline
%Median   & \textbf{21.344} & 21.425 & 23.953 \\ \hline
%Minimum   & \textbf{18.572} & 20.753 & 21.8896 \\ \hline 
%\end{tabular}
%\end{center}

\begin{center}
\begin{tabular}{| c |c c c|}
\multicolumn{1}{c}{Method} & \multicolumn{3}{c}{Validation PPL} \\  
\hline
                               & Min & Median & Max \\ \hline
                          Adam & 20.84 & 21.83 & 21.86 \\ \hline
                    AdaHessian & 22.12 & 24.23 & 25.79 \\ \hline
                 Alice, $\rho$ & 21.07 & 23.43 & 24.95 \\ \hline
          Alice, $h_{\rm abs}$ & \textbf{18.66} & \textbf{21.50} & 23.74 \\ \hline
\end{tabular}
\end{center}

}

\extended % !!!PUBLISH-CHECK!!! Comment to remove extended material

% =================================================================
% ======================== Acknowledgments ========================
% =================================================================
\section*{Acknowledgments}
\anonymize[Anonymized.]{%
Sandia National Laboratories is a multimission laboratory managed and operated by National Technology and Engineering Solutions of Sandia, LLC.,
a wholly owned subsidiary of Honeywell International, Inc., for the U.S. Department of Energy's National Nuclear Security Administration under contract
DE-NA-0003525. This paper describes objective technical results and analysis. Any subjective views or opinions that might be expressed in the paper
do not necessarily represent the views of the U.S. Department of Energy or the United States Government. \\
\\
We appreciate feedback during writing from Esha Datta, Connor Mattes, and Arvind Prasadan that helped improve the clarity of this work.
}

% ============================================================
% ======================== References ========================
% ============================================================
\bibliography{main_112524.bib}

\newcommand{\reproducibility}{

\section*{Reproducibility Checklist}

\begin{itemize}
\item Includes a conceptual outline and/or pseudocode description of AI methods introduced: Yes.
\item Clearly delineates statements that are opinions, hypothesis, and speculation from objective facts and results: Yes.
\item Provides well marked pedagogical references for less-familiar readers to gain background necessary to replicate the paper: Yes.
\item Does this paper make theoretical contributions? Yes.
\item All assumptions and restrictions are stated clearly and formally: Yes.
\item All novel claims are stated formally (e.g., in theorem statements): Yes.
\item Proofs of all novel claims are included: Yes.
\item Proof sketches or intuitions are given for complex and/or novel results: Partial. We do not expand upon the optimal sample kernel beyond the proof.
\item Appropriate citations to theoretical tools used are given: Yes.
\item All theoretical claims are demonstrated empirically to hold: Partial. We provide experimental evidence that these methods can improve optimization.
\item All experimental code used to eliminate or disprove claims is included: Partial. We are awaiting approval to release our source code, but the optimization data that we generated and used in figures and summary statistics is included. We aim to make source code available before peer review is completed.
\item Does this paper rely on one or more datasets? No. While we use CIFAR-10, ImageNet, and WikiText-103 datasets to demonstrate optimization comparisons, our key results are theoretical and do not rely on these demonstrations.

\item A motivation is given for why the experiments are conducted on the selected datasets: No.
\item All novel datasets introduced in this paper are included in a data appendix: NA.
\item All novel datasets introduced in this paper will be made publicly available upon publication of the paper with a license that allows free usage for research purposes: NA
\item All datasets drawn from the existing literature (potentially including authors’ own previously published work) are accompanied by appropriate citations: Yes.
\item All datasets drawn from the existing literature (potentially including authors’ own previously published work) are publicly available: Yes.
\item All datasets that are not publicly available are described in detail, with explanation why publicly available alternatives are not scientifically satisfying: NA.
\item Does this paper include computational experiments? Yes.
\item Any code required for pre-processing data is included in the appendix: Yes.
\item All source code required for conducting and analyzing the experiments is included in a code appendix: Yes.
\item All source code required for conducting and analyzing the experiments will be made publicly available upon publication of the paper with a license that allows free usage for research purposes: Yes.
\item All source code implementing new methods have comments detailing the implementation, with references to the paper where each step comes from: Yes.
\item If an algorithm depends on randomness, then the method used for setting seeds is described in a way sufficient to allow replication of results: Yes.
\item This paper specifies the computing infrastructure used for running experiments (hardware and software), including GPU/CPU models; amount of memory; operating system; names and versions of relevant software libraries and frameworks: Yes.
\item This paper formally describes evaluation metrics used and explains the motivation for choosing these metrics: Yes.
\item This paper states the number of algorithm runs used to compute each reported result: Yes.
\item Analysis of experiments goes beyond single-dimensional summaries of performance (e.g., average; median) to include measures of variation, confidence, or other distributional information: Yes.
\item The significance of any improvement or decrease in performance is judged using appropriate statistical tests: No. The visual results are sufficient to show that improvements are possible with these methods.
\item This paper lists all final (hyper-)parameters used for each model/algorithm in the paper’s experiments: Yes.
\item This paper states the number and range of values tried per (hyper-) parameter during development of the paper, along with the criterion used for selecting the final parameter setting: Partial. 
\end{itemize}

}

% \reproducibility % !!!PUBLISH-CHECK!!! Comment to remove reproducibility checklist

\end{document}